\documentclass{article}

\usepackage{makeidx}  % allows for indexgeneration

\usepackage{graphicx}
\usepackage{cite}

\usepackage{color}
\usepackage{amsmath}
\usepackage{epsfig}
\usepackage{url}

\usepackage{bm}

\newtheorem{thm}{Theorem}
\newtheorem{lem}{Lemma}
\newtheorem{corl}{Corollary}

\usepackage{amssymb}
\usepackage{amsfonts}

\newcommand{\qed}{\fbox{\rule[1pt]{0pt}{0pt}}}
\newcommand{\bsquare}{\hbox{\rule{6pt}{6pt}}}

\newcommand{\bfb}{{\bf b}}

\newcommand{\bfe}{{\bf e}}
\newcommand{\bfx}{{\bf x}}

\newcommand{\bfr}{{\bf r}}

\newcommand{\Wp}{{\widetilde{\bf W}}}

\newcommand{\Be}{{\widehat{\bf B}}}
\newcommand{\Bp}{{\widetilde{\bf B}}}

\newcommand{\R}{{\bf R}}
\newcommand{\A}{{\bf A}}
\newcommand{\B}{{\bf B}}

\newcommand{\W}{{\bf W}}
\newcommand{\I}{{\bf I}}
\newcommand{\D}{{\bf D}}

\newcommand{\x}{{\bf x}}
\newcommand{\X}{{\bf X}}
\newcommand{\bP}{{\bf P}}

\title{DirectLiNGAM: A direct method for learning a linear non-Gaussian structural equation model}
\author{Shohei Shimizu\thanks{The Institute of Scientific and Industrial Research (ISIR), Osaka University, Mihogaoka 8-1, Ibaraki, Osaka 567-0047, Japan. Email: sshimizu@ar.sanken.osaka-u.ac.jp}, 
Takanori Inazumi\thanks{The Institute of Scientific and Industrial Research (ISIR), Osaka University, Mihogaoka 8-1, Ibaraki, Osaka 567-0047, Japan. Email: inazumi@ar.sanken.osaka-u.ac.jp}, 
Yasuhiro Sogawa\thanks{The Institute of Scientific and Industrial Research (ISIR), Osaka University, Mihogaoka 8-1, Ibaraki, Osaka 567-0047, Japan. Email: sogawa@ar.sanken.osaka-u.ac.jp}, \\
Aapo Hyv\"arinen\thanks{Department of Computer Science, Department of Mathematics and Statistics and Helsinki Institute for Information Technology, University of Helsinki, FIN-00014, Finland. Email: aapo.hyvarinen@helsinki.fi}, 
Yoshinobu Kawahara\thanks{The Institute of Scientific and Industrial Research (ISIR), Osaka University, Mihogaoka 8-1, Ibaraki, Osaka 567-0047, Japan. Email: kawahara@ar.sanken.osaka-u.ac.jp}, 
Takashi Washio\thanks{The Institute of Scientific and Industrial Research (ISIR), Osaka University, Mihogaoka 8-1, Ibaraki, Osaka 567-0047, Japan. Email: washio@ar.sanken.osaka-u.ac.jp}\\
Patrik O. Hoyer\thanks{Department of Computer Science, University of Helsinki, FIN-00014, Finland. Email: patrik.hoyer@helsinki.fi}, 
Kenneth Bollen\thanks{Department of Sociology, CB 3210 Hamilton Hall, University of North Carolina, Chapel Hill, NC 27599-3210, U.S.A. Email: bollen@unc.edu}, 
}
\date{}

\begin{document}

\maketitle

\begin{abstract}
Structural equation models and Bayesian networks have been widely used to analyze causal relations between continuous variables. In such frameworks, linear acyclic models are typically used to model the data-generating process of variables. 
Recently, it was shown that use of non-Gaussianity identifies the full structure of a linear acyclic model, {\it i.e.}, a causal ordering of variables and their connection strengths, without using any prior knowledge on the network structure, which is not the case with conventional methods. 
However, existing estimation methods are based on iterative search algorithms and may not converge to a correct solution in a finite number of steps. 
In this paper, we propose a new direct method to estimate a causal ordering and connection strengths based on non-Gaussianity. 
 In contrast to the previous methods, our algorithm requires no algorithmic parameters and is guaranteed to converge to the right solution within a small fixed number of steps if the data strictly follows the model. 
 \end{abstract}

%\newpage
%\tableofcontents
%\newpage

%%%%%%%%%%%%%%%%%%%%%%%%%%%%%%%%%%%%
\section{Introduction}\label{sec:intro}
%%%%%%%%%%%%%%%%%%%%%%%%%%%%%%%%%%%%
Many empirical sciences aim to discover and understand causal mechanisms underlying various natural phenomena and human social behavior. 
An effective way to study causal relationships is to conduct a controlled experiment. 
However, performing controlled experiments is often ethically impossible or too expensive in many fields including social sciences \cite{Bollen89book}, bioinformatics \cite{Rhein07BMCSB} and neuroinformatics \cite{Londei06CP}. 
Thus, it is necessary and important to develop methods for causal inference based on the data that do not come from such controlled experiments. 

Structural equation models (SEM) \cite{Bollen89book} and Bayesian networks (BN) \cite{Pearl00book,Spirtes93book} are widely applied to analyze causal relationships in many empirical studies. 
A \emph{linear} acyclic model that is a special case of SEM and BN is typically used to analyze causal effects between continuous variables.  
Estimation of the model commonly uses only the covariance structure of the data and in most cases cannot identify the full structure, {\it i.e.}, a causal ordering and connection strengths, of the model with no prior knowledge on the structure \cite{Pearl00book,Spirtes93book}. 

In \cite{Shimizu06JMLR}, a non-Gaussian variant of SEM and BN called a linear non-Gaussian acyclic model (LiNGAM) was proposed, and its full structure was shown to be identifiable without pre-specifying a causal order of the variables. 
This feature is a significant advantage over the conventional methods \cite{Spirtes93book,Pearl00book}.  
A non-Gaussian method to estimate the new model was also developed in \cite{Shimizu06JMLR} 
and is closely related to independent component analysis (ICA) \cite{Hyva01book}. 
In the subsequent studies, the non-Gaussian framework has been extended in various directions for learning a wider variety of SEM and BN \cite{Hoyer09NIPS,Hyva10JMLR,Lacerda08UAI}. 
In what follows, we refer to the non-Gaussian model as LiNGAM and the estimation method as ICA-LiNGAM algorithm. 

Most of major ICA algorithms including \cite{Amari-Natural-Gradient,Hyva99TNN} are iterative search  methods \cite{Hyva01book}. 
Therefore, the ICA-LiNGAM algorithms based on the ICA algorithms need some additional information including initial guess and convergence criteria. Gradient-based methods \cite{Amari-Natural-Gradient} further need step sizes. 
However, such algorithmic parameters are hard to optimize in a systematic way. 
Thus, the ICA-based algorithms may get stuck in local optima and may not converge to a reasonable solution if the initial guess is badly chosen \cite{Himberg04icasso}.

In this paper, we propose a new direct method to estimate a causal ordering of variables in the LiNGAM without prior knowledge on the structure. 
The new method estimates a causal order of variables by successively reducing each independent component from given data in the model, and this process is completed in steps equal to the number of the variables in the model. 
It is not based on iterative search in the {\it parameter space} and needs no initial guess or similar algorithmic parameters. 
It is {\it guaranteed} to converge to the right solution within a small fixed number of steps if the data {\it strictly} follows the model, {\it i.e.}, if all the model assumptions are met and the sample size is infinite. 
These features of the new method enable more accurate estimation of a causal order of the variables in a disambiguated and direct procedure.
Once the causal orders of variables is identified, the connection strengths between the variables are easily estimated using 
some conventional covariance-based methods such as least squares and maximum likelihood approaches \cite{Bollen89book}. 
We also show how prior knowledge on the structure can be incorporated in the new method. 

The paper is structured as follows. 
First, in Section~\ref{sec:background}, we briefly review LiNGAM and the ICA-based LiNGAM algorithm. 
We then in Section~\ref{sec:dir} introduce a new direct method. 
The performance of the new method is examined by experiments on artificial data in Section~\ref{sec:exp}, and experiments on real-world data in Section~\ref{sec:real}. 
Conclusions are given in Section~\ref{sec:conc}.
Preliminary results were presented in \cite{Shimizu09UAI,Inazumi10LVA,Sogawa10IJCNN}. 

%%%%%%%%%%%%%%%%%%%%%%%%%%%%%%%%%%%%
\section{Background}\label{sec:background}
%%%%%%%%%%%%%%%%%%%%%%%%%%%%%%%%%%%%

%%%%%%%%%%%%%%%%%%%%%%%%%%%%%%%%%%%%
\subsection{A linear non-Gaussian acyclic model: LiNGAM}\label{sec:model}
%%%%%%%%%%%%%%%%%%%%%%%%%%%%%%%%%%%%
In \cite{Shimizu06JMLR}, a non-Gaussian variant of SEM and BN, which is called LiNGAM, was proposed. 
Assume that observed data are generated from a process represented graphically by a directed acyclic graph, {\it i.e.}, DAG. 
Let us represent this DAG by a $m$$\times$$m$  adjacency matrix $\B$$=$$\{b_{ij}\}$ where every $b_{ij}$ represents the connection strength from a variable $x_j$ to another $x_i$ in the DAG. 
Moreover, let us denote by $k(i)$ a causal order of variables $x_i$ in the DAG so that no later variable determines or has a directed path on any earlier variable. (A directed path from $x_i$ to $x_j$ is a sequence of directed edges such that $x_j$ is reachable from $x_i$.)
We further assume that the relations between variables are linear. 
Without loss of generality, each observed variable $x_i$ is assumed to have zero mean. 
Then we have
\begin{equation}
x_i = \sum_{k(j)<k(i)} b_{ij}x_j + e_i,\label{eq:model0}
\end{equation}
where $e_i$ is an external influence. 
All external influences $e_i$ are continuous random variables having \emph{non-Gaussian} distributions with zero means and non-zero variances, and $e_i$ are independent of each other so that there is no latent confounding variables \cite{Spirtes93book}. 

We rewrite the model (\ref{eq:model0}) in a matrix form as follows: 
\begin{equation}
\bfx = \B\bfx + \bfe,\label{eq:model}
\end{equation} 
where $\bfx$ is a $p$-dimensional random vector, and $\B$ could be permuted by simultaneous equal row and column permutations to be {\it strictly} lower triangular due to the acyclicity assumption \cite{Bollen89book}. Strict lower triangularity is here defined as a lower triangular structure with all zeros on the diagonal. 
Our goal is to estimate the adjacency matrix $\B$ by observing data $\bfx$ only. 
Note that we do not assume that the distribution of $\bfx$ is faithful \cite{Spirtes93book} to the generating graph. 

We note that each $b_{ij}$ represents the direct causal effect of $x_j$ on $x_i$ and each $a_{ij}$, the $(i,j)$-the element of the matrix $\A$$=$$(\I-\B)^{-1}$, the total causal effect of $x_j$ on $x_i$ \cite{Hoyer07IJAR}. 

We emphasize that $x_i$ is equal to $e_i$ if no other observed variable $x_j$ ($j$$\neq$$i$) inside the model has a directed edge to $x_i$, {\it i.e.}, all the $b_{ij}$ ($j$$\neq$$i$) are zeros. 
In such a case, an external influence $e_i$ is {\it observed} as $x_i$. 
Such an $x_i$ is called an {\it exogenous observed} variable. Otherwise, $e_i$ is called an {\it error}. 
For example, consider the model defined by
\begin{eqnarray*}
x_2 &=& e_2 \\
x_1 &=& 1.5 x_2 + e_1 \\
x_3 &=& 0.8 x_1 -1.5  x_2 + e_3, 
\end{eqnarray*}
where $x_2$ is equal to $e_2$ since it is not determined by either $x_1$ or $x_3$. 
Thus, $x_2$ is an exogenous observed variable, and $e_1$ and $e_3$ are errors. 
Note that there {\it exists at least one exogenous observed variable} $x_i$($=$$e_i$) due to the acyclicity and the assumption of no latent confounders. 

An exogenous observed variable is usually defined as an observed variable that is determined outside of the model \cite{Bollen89book}. 
In other words, an exogenous observed variable is a variable that any other observed variable inside the model does not have a directed edge to. The definition does not require that it is equal to an independent external influence, and the external influences of exogenous observed variables may be dependent. However, in the LiNGAM (\ref{eq:model}), an exogenous observed variable is always equal to an independent external influence due to the assumption of no latent confounders.

%%%%%%%%%%%%%%%%%%%%%%%%%%%%%%%%%%%%
\subsection{Identifiability of the model}\label{sec:identifiablity}
%%%%%%%%%%%%%%%%%%%%%%%%%%%%%%%%%%%%
We next explain how the connection strengths of the LiNGAM (\ref{eq:model}) can be identified as shown in \cite{Shimizu06JMLR}. Let us first solve Eq.~(\ref{eq:model}) for $\bfx$. Then we obtain
\begin{equation} \label{eq:ica}
\bfx = \A\bfe,
\end{equation}
where $\A= (\I-\B)^{-1}$ is a mixing matrix whose elements are called mixing coefficients and can be permuted to be lower triangular as well due to the aforementioned feature of $\B$ and the nature of matrix inversion.
Since the components of $\bfe$ are independent and non-Gaussian, Eq.~(\ref{eq:ica}) defines the  independent component analysis (ICA) model \cite{Hyva01book}, which is known to be identifiable \cite{Comon94,Erik04SPL}.

ICA essentially can estimate $\A$ (and $\W=\A^{-1}=\I-\B$), but has permutation, scaling and sign
 indeterminacies. 
ICA actually gives $\W_{ICA}$$=$$\bP\D\W$, where $\bP$ is an unknown permutation matrix, and $\D$ is an unknown diagonal matrix. 
But in LiNGAM, the correct permutation matrix $\bP$ can be found \cite{Shimizu06JMLR}: the correct $\bP$ is the only one that gives no zeros in the diagonal of $\D\W$ since $\B$ should be a matrix that can be permuted to be strictly lower triangular and $\W=\I-\B$.  
Further, one can find the correct scaling and signs of the independent components by using the unity on the diagonal of $\W$$=$$\I$$-$$\B$. 
One only has to divide the rows of $\D\W$ by its corresponding diagonal elements to obtain $\W$.
Finally, one can compute the connection strength matrix  $\B=\I-\W$.

%%%%%%%%%%%%%%%%%%%%%%%%%%%%%%%%%%%%
\subsection{ICA-LiNGAM algorithm}\label{sec:prev}
%%%%%%%%%%%%%%%%%%%%%%%%%%%%%%%%%%%%
The ICA-LiNGAM algorithm presented in \cite{Shimizu06JMLR} is described as follows: 

\noindent
  %\pagebreak
  \rule{\columnwidth}{0.5mm}
       { \sffamily
	ICA-LiNGAM algorithm
	 \begin{enumerate}
	 \item \label{step:ica} Given a $p$-dimensional random vector $\x$ and its $p \times n$ observed data matrix $\X$, apply an ICA algorithm (FastICA using hyperbolic tangent function \cite{Hyva99TNN}) to obtain an estimate of $\A$. 
	 \item \label{step:rowperm} Find the unique permutation of rows of $\W$$=$$\A^{-1}$ which yields a matrix $\Wp$ without any zeros on the main diagonal. 
	 The permutation is sought by minimizing $\sum_i 1/|{\Wp_{ii}}|$.
	 \item \label{step:normalize} Divide each row of $\Wp$ by its corresponding diagonal element, to yield a new matrix $\Wp'$ with all ones on the diagonal.
	 \item \label{step:computeB} Compute an estimate $\Be$ of $\B$ using $\Be = \I - \Wp'$.
	 \item \label{step:causalperm} Finally, to estimate a causal order $k(i)$, find the permutation matrix $\widetilde{\bP}$  of $\Be$ yielding a matrix $\Bp = \widetilde{\bP} \Be \widetilde{\bP}^T$ which is as close as possible to a strictly lower triangular structure. The lower-triangularity of $\Bp$ can be measured using the sum of squared $b_{ij}$ in its upper triangular part $\sum_{i\leq j} \widetilde{b}_{ij}^2$ for small number of variables, say less than 8. For higher-dimensional data, the following approximate algorithm is used, which sets small absolute valued elements in $\Bp$ to zero and tests if the resulting matrix is possible to be permuted to be strictly lower triangular: 
\begin{enumerate}
\item Set the $p(p+1)/2$ smallest (in absolute value) elements of $\Be$ to zero. 
\item Repeat
\begin{enumerate}
\item Test if $\Be$ can be permuted to be strictly lower triangular. If the answer is yes, stop and return the permuted $\Be$, that is, $\Bp$. 
\item Additionally set the next smallest (in absolute value) element of $\Be$ to zero. 
\end{enumerate}
\end{enumerate}
	 \end{enumerate}
       } \vspace{-4mm}
\noindent \rule{\columnwidth}{0.5mm}

\subsection{Potential problems of ICA-LiNGAM}
The original ICA-LiNGAM algorithm has several potential problems: i) Most ICA algorithms including FastICA \cite{Hyva99TNN} and gradient-based algorithms \cite{Amari-Natural-Gradient} may not converge to a correct solution in a finite number of steps if the initially guessed state is badly chosen \cite{Himberg04icasso} or if the step size is not suitably selected for those gradient-based methods.
The appropriate selection of such algorithmic parameters is not easy. 
In contrast, our algorithm proposed in the next section is guaranteed to converge to the right solution in a fixed number of steps equal to the number of variables if the data {\it strictly} follows the model. 
ii) The permutation algorithms in Steps~\ref{step:rowperm} and~\ref{step:causalperm} are not scale-invariant. Hence they could give a different or {\it even wrong} ordering of variables depending on scales or standard deviations of variables especially when they have a wide range of scales. However, scales are essentially not relevant to the ordering of variables. 
Though such bias would vanish for large enough sample sizes, for practical sample sizes, an estimated ordering could be affected when variables are normalized to make unit variance for example, and hence the estimation of a causal ordering becomes quite difficult. 

%%%%%%%%%%%%%%%%%%%%%%%%%%%%%%%%%%%%
\section{A direct method: DirectLiNGAM}\label{sec:dir}
%\section{A direct method: DirectLiNGAM}\label{sec:dir}
%%%%%%%%%%%%%%%%%%%%%%%%%%%%%%%%%%%%

\subsection{Identification of an exogenous variable based on non-Gaussianity and independence}
In this subsection, we present two lemmas and a corollary\footnote{We prove the lemmas and corollary without assuming the faithfulness \cite{Spirtes93book} unlike our previous work \cite{Shimizu09UAI}.} that ensure the validity of our algorithm proposed in the next subsection~\ref{sec:alg}. 
The basic idea of our method is as follows. 
We first find an exogenous variable based on its independence of the residuals of a number of pairwise regressions (Lemma~\ref{lemma1}). 
Next, we remove the effect of the exogenous variable from the other variables using least squares regression. 
Then, we show that a LiNGAM also holds for the residuals (Lemma~\ref{lemma2}) and that the same ordering of the residuals is a causal ordering for the original observed variables as well (Corollary~\ref{corollary3}). 
Therefore, we can find the second variable in the causal ordering of the original observed variables by analyzing the residuals and their LiNGAM, {\it i.e.}, by applying Lemma~\ref{lemma1} to the residuals and finding an ``exogenous'' residual.  
The iteration of these effect removal and causal ordering estimates the causal order of the original variables.

We first quote Darmois-Skitovitch theorem \cite{Darmois1953,Skitovitch53} since it is used to prove Lemma~\ref{lemma1}:
\begin{thm}[Darmois-Skitovitch theorem]\label{thm1}
Define two random variables $y_1$ and $y_2$ as linear combinations of independent random variables $s_i$($i$$=$1, $\cdots$, $q$): 
\begin{eqnarray}
y_1 = \sum_{i=1}^q \alpha_is_i, \hspace{2mm}y_2 = \sum_{i=1}^q \beta_i s_i.
\end{eqnarray}
Then, if $y_1$ and $y_2$ are independent, all variables $s_j$ for which $\alpha_j\beta_j\neq0$ are Gaussian. \mbox{\hfill \qed}
\end{thm}
In other words, this theorem means that if there exists a non-Gaussian $s_j$ for which $\alpha_j\beta_j$$\neq$$0$, $y_1$ and $y_2$ are dependent. 

\begin{lem}\label{lemma1}
Assume that the input data $\bfx$ strictly follows the LiNGAM (\ref{eq:model}). 
Denote by $r_i^{(j)}$ the residuals  when $x_i$ are regressed on $x_j$: 
$r_i^{(j)} = x_i - \frac{{\rm cov}(x_i,x_j)}{{\rm var}(x_j)}x_j$ $(i \neq j)$.
Then a variable $x_j$ is exogenous if and only if $x_j$ is independent of its residuals $r_i^{(j)}$ for all $i \neq j$. \mbox{\hfill \qed}
\end{lem} 

\paragraph{Proof}
(i) Assume that $x_j$ is exogenous, {\it i.e.}, $x_j$$=$$e_j$. 
Due to the model assumption and Eq.~(\ref{eq:ica}), one can write 
$x_i$$=$$a_{ij}$$x_j$$+$$\bar{e}_i^{(j)}$ ($i$$\neq$$j$), where $\bar{e}_i^{(j)}$$=$$\sum_{h\neq j} a_{ih}e_h$ and $x_j$ are independent, and $a_{ij}$ is a mixing coefficient from $x_j$ to $x_i$ in Eq.~(\ref{eq:ica}). The mixing coefficient $a_{ij}$ is equal to the regression coefficient when $x_i$ is regressed on $x_j$ since ${\rm cov}(x_i,x_j)$$=$$a_{ij}$${\rm var}(x_j)$. 
Thus,  the residual $r_i^{(j)}$ is equal to the corresponding error term, {\it i.e.}, $r_i^{(j)}$$=$$\bar{e}_i^{(j)}$. This implies that $x_j$ and $r_i^{(j)}$$($$=$$\bar{e}_i^{(j)}$$)$ are independent.

(ii) Assume that $x_j$ is not exogenous, {\it i.e.}, $x_j$ has at least one parent. 
Let $P_j$ denote the (non-empty) set of the variable subscripts of parent variables of $x_j$. 
Then one can write $x_j = \sum_{h \in P_j} b_{jh} x_h + e_j$, where $x_h$ and $e_j$ are independent and each $b_{jh}$ is non-zero. 
Let a vector $\x_{P_j}$ and a column vector $\bfb_{P_j}$ collect all the variables in $P_j$ and the corresponding connection strengths, respectively. 
Then, the covariances between $\x_{P_j}$ and $x_j$ are 
\begin{eqnarray}
E(\x_{P_j}x_j) &=& E\{\x_{P_j} (\bfb_{P_j}^T \x_{P_j}+e_j) \} \nonumber\\
 &=& E(\x_{P_j}\bfb_{P_j}^T \x_{P_j})+E( \x_{P_j}e_j) \nonumber \\
 &=& E(\x_{P_j} \x_{P_j}^T)\bfb_{P_j}. \label{eq:cov}
\end{eqnarray}
The covariance matrix $E(\x_{P_j} \x_{P_j}^T)$ is positive definite since the external influences $e_h$ that correspond to those parent variables $x_h$ in $P_j$ are mutually independent and have  positive variances.
%Those parent variables $x_h$ in $P_j$ are not collinear since the corresponding external influences $e_h$ are independent and have  positive variances, which implies that their covariance matrix $E(\x_{P_j} \x_{P_j}^T)$ is positive definite.
Thus, the covariance vector $E(\x_{P_j}x_j)=E(\x_{P_j} \x_{P_j}^T)\bfb_{P_j}$ in Eq.~(\ref{eq:cov}) cannot equal the zero vector, 
and there must be at least one variable $x_i$ ($i \in P_j$) with which $x_j$ covaries, {\it i.e.}, {\rm cov}$(x_i,x_j)$$\neq$$0$. 
Then, for such a variable $x_i$ ($i \in P_j$) that {\rm cov}$(x_i,x_j)$$\neq$$0$, we have 
\begin{eqnarray}
r_i^{(j)} &=& x_i - \frac{{\rm cov}( x_i, x_j )}{{\rm var}(x_j)}x_j \nonumber \\
 &=& x_i - \frac{{\rm cov}( x_i, x_j )}{{\rm var}(x_j)}\left(\sum_{h \in P_j} b_{jh} x_h + e_j\right) \nonumber \\
 &=& \left\{1-\frac{b_{ji}{\rm cov}(x_i,x_j)}{{\rm var}(x_j)}\right\} x_i -\frac{{\rm cov}(x_i,x_j)}{{\rm var}(x_j)}\sum_{h\in P_j,  h\neq i}b_{jh}x_h \nonumber \\
  & & - \frac{{\rm cov}(x_i,x_j)}{{\rm var}(x_j)} e_j. 
% &=& \left\{1-\frac{b_{ji}{\rm cov}(x_i,x_j)}{{\rm var}(x_j)}\right\} x_i - \frac{{\rm cov}(x_i,x_j)}{{\rm var}(x_j)} e_j \nonumber \\
%  & & -\frac{{\rm cov}(x_i,x_j)}{{\rm var}(x_j)}\sum_{x_h\in P_j, h\neq i}b_{jh}x_h. \\
\end{eqnarray}
Each of those parent variables $x_h$ (including $x_i$) in $P_j$ is a linear combination of external influences {\it other than} $e_j$ due to the relation of $x_h$ to $e_j$ that $x_j = \sum_{h \in P_j} b_{jh} x_h + e_j = \sum_{h \in P_j} b_{jh} \left(\sum_{k(t)\le k(h)} a_{ht}e_t\right) + e_j$ , where $e_t$ and $e_j$ are independent. 
%Each of those parent variables $x_i$ and $x_h$ in $P_j$ is a linear combination of external influences other than $e_j$ due to Eq.~(\ref{eq:ica}) and the equation $x_j = \sum_{h \in P_j} b_{jh} x_h + e_j$. %, where $x_h$ and $e_j$ are independent. 
Thus, the $r_i^{(j)}$ and $x_j$ can be rewritten as linear combinations of independent external influences as follows: 
%Since each of those parent variables $x_i$ and $x_h$ in $P_j$ is a linear combination of external influences other than $e_j$, the $r_i^{(j)}$ and $x_j$ can be written as linear combinations of independent external influences: 
\begin{eqnarray}
r_i^{(j)}  
%&=& \left\{1-\frac{b_{ji}{\rm cov}(x_i,x_j)}{{\rm var}(x_j)}\right\} x_i -\frac{{\rm cov}(x_i,x_j)}{{\rm var}(x_j)}\sum_{h\in P_j,  h\neq i}b_{jh}x_h \nonumber \\
%  & & - \frac{{\rm cov}(x_i,x_j)}{{\rm var}(x_j)} e_j \nonumber \\
&=& \left\{1-\frac{b_{ji}{\rm cov}(x_i,x_j)}{{\rm var}(x_j)}\right\} \left(\sum_{l\neq j} a_{il}e_l\right) -\frac{{\rm cov}(x_i,x_j)}{{\rm var}(x_j)}\sum_{h\in P_j, h\neq i}b_{jh}\left(\sum_{t\neq j} a_{ht}e_t\right) \nonumber \\
% &=& \left\{1-\frac{b_{ji}{\rm cov}(x_i,x_j)}{{\rm var}(x_j)}\right\} \left(\sum_{l\neq j} a_{il}e_l\right) -\frac{{\rm cov}(x_i,x_j)}{{\rm var}(x_j)}\sum_{x_h\in P_j, h\neq i}b_{jh}\left(\sum_{q\neq j} a_{hq}e_q\right) \nonumber \\  
& & - \frac{{\rm cov}(x_i,x_j)}{{\rm var}(x_j)} e_j. \label{eq:r}\\
x_j 
%&=& \sum_{h \in P_j} b_{jh} x_h + e_j \nonumber \\
 &=& \sum_{h \in P_j} b_{jh} \left(\sum_{t\neq j} a_{ht}e_t\right) + e_j.\label{eq:xj} 
\end{eqnarray}
The first two terms of Eq.~(\ref{eq:r}) and the first term of Eq.~(\ref{eq:xj}) are linear combinations of external influences other than $e_j$, and the third term of Eq.~(\ref{eq:r}) and the second term of Eq.~(\ref{eq:xj}) depend only on $e_j$ and do not depend on the other external influences. 
Further, all the external influences including $e_j$ are mutually independent, and the coefficient of non-Gaussian $e_j$ on $r_i^{(j)}$ and that on $x_j$ are non-zero. 
%The coefficient of $e_j$ on $r_i^{(j)}$ and that on $x_j$ are non-zero and $e_j$ is non-Gaussian and independent of all $x_h, x_i$$\in$$P_j$. 
These imply that $r_i^{(j)}$ and $x_j$ are dependent since $r_i^{(j)}$, $x_j$ and $e_j$ correspond to $y_1$, $y_2$, $s_j$ in Darmois-Skitovitch theorem, respectively. 

From (i) and (ii), the lemma is proven.  \mbox{\hfill \bsquare}

\begin{lem}\label{lemma2}
%Assume the assumptions in Lemma~\ref{lemma1}. 
Assume that the input data $\bfx$ strictly follows the LiNGAM (\ref{eq:model}). 
Further, assume that a variable $x_j$ is exogenous. 
Denote by $\bfr^{(j)}$ a ($p$-1)-dimensional vector that collects the residuals $r_i^{(j)}$ when all $x_i$ of $\bfx$ are regressed on $x_j$ $($$i$$\neq$$j$$)$. 
Then a LiNGAM holds for the residual vector $\bfr^{(j)}$: 
$\bfr^{(j)}=\B^{(j)}\bfr^{(j)} + \bfe^{(j)}$, where $\B^{(j)}$ is a matrix that can be permuted to be strictly lower-triangular by a simultaneous row and column permutation, and elements of $\bfe^{(j)}$ are non-Gaussian and mutually independent. \mbox{\hfill \qed}
\end{lem} 

\paragraph{Proof}
Without loss of generality, assume that $\B$ in the LiNGAM (\ref{eq:model}) is already permuted to be strictly lower triangular and that $x_j$$=$$x_1$. 
Note that $\A$ in Eq.~(\ref{eq:ica}) is also lower triangular (although its diagonal elements are all ones).
Since $x_1$ is exogenous, $a_{i1}$ are equal to the regression coefficients when $x_i$ are regressed on $x_1$ $(i \neq 1)$.  
Therefore, after removing the effects of $x_1$ from $x_i$ by least squares estimation, one gets the first column of $\A$ to be a zero vector, and $x_1$ does not affect the residuals $r_i^{(1)}$. 
Thus, we again obtain a lower triangular mixing matrix $\A^{(1)}$ with all ones in the diagonal for the residual vector $\bfr^{(1)}$ and hence have a LiNGAM for the vector $\bfr^{(1)}$. \mbox{\hfill \bsquare}

\begin{corl}\label{corollary3}
%Assume the assumptions in Lemma~\ref{lemma2}. 
Assume that the input data $\bfx$ strictly follows the LiNGAM (\ref{eq:model}). 
Further, assume that a variable $x_j$ is exogenous. 
Denote by $k_{r^{(j)}}(i)$ a causal order of $r_i^{(j)}$. 
Recall that  $k(i)$ denotes a causal order of $x_i$.  
Then, the same ordering of the residuals is a causal ordering for the original observed variables as well: 
$k_{r^{(j)}}(l)$$<$$k_{r^{(j)}}(m)$ $\Leftrightarrow$ $k(l)$$<$$k(m)$. \mbox{\hfill \qed}
\end{corl} 

\paragraph{Proof}
As shown in the proof of Lemma~\ref{lemma2}, when the effect of an exogenous variable $x_1$ is removed from the other observed variables, the second to $p$-th columns of $\A$ remain the same, and the submatrix of $\A$ formed by deleting the first row and the first column is still lower triangular. 
This shows that the ordering of the other variables is not changed and proves the corollary.  
 \mbox{\hfill \bsquare}

Lemma~\ref{lemma2} indicates that the LiNGAM for the ($p$$-$$1$)-dimensional residual vector $\bfr^{(j)}$ can be handled as a new input model, and Lemma~\ref{lemma1} can be further applied to the model to estimate the next exogenous variable (the next exogenous residual in fact). This process can be repeated until all variables are ordered, and the resulting order of the variable subscripts shows the causal order of the original observed variables according to Corollary~\ref{corollary3}.

To apply Lemma~\ref{lemma1} in practice, we need to use a measure of independence which is not restricted to uncorrelatedness since least squares regression gives residuals always uncorrelated with but not necessarily independent of explanatory variables. 
A common independence measure between two variables $y_1$ and $y_2$ is their mutual information $MI(y_1,y_2)$ \cite{Hyva01book}. In \cite{Bach02JMLR}, a nonparametric estimator of mutual information was developed using kernel methods.\footnote{Matlab codes can be downloaded at \url{http://www.di.ens.fr/~fbach/kernel-ica/index.htm}}  
Let $K_1$ and $K_2$ represent the Gram matrices whose elements are Gaussian kernel values of the sets of $n$ observations of $y_1$ and $y_2$, respectively. 
The Gaussian kernel values $K_1(y_1^{(i)},y_1^{(j)})$ and $K_2(y_2^{(i)},y_2^{(j)})$  $(i,j=1,\cdots,n)$ are computed by
\begin{eqnarray}
K_1(y_1^{(i)},y_1^{(j)}) &=& \exp \left( -\frac{1}{2\sigma^2}\|y_1^{(i)}-y_1^{(j)}\|^2\right)\\
K_2(y_2^{(i)},y_2^{(j)}) &=& \exp \left( -\frac{1}{2\sigma^2}\|y_2^{(i)}-y_2^{(j)}\|^2\right), 
\end{eqnarray}
where $\sigma$$>$$0$ is the bandwidth of Gaussian kernel. 
Further let $\kappa$ denote a small positive constant. 
Then, in \cite{Bach02JMLR}, the kernel-based estimator of mutual information is defined as: 
\begin{eqnarray}
\widehat{MI}_{kernel}(y_1,y_2) = -\frac{1}{2} \log \frac{\det \mathcal{K}_{\kappa}}{\det \mathcal{D}_{\kappa}},
\end{eqnarray}
where 
\begin{eqnarray}
\mathcal{K}_{\kappa} &=& 
\left[
\begin{array}{cc}
\left( K_1 + \frac{n\kappa}{2}I\right)^2 & K_1K_2\\
K_2K_1 & \left( K_2 + \frac{n\kappa}{2}I\right)^2
\end{array}
\right]\\
\mathcal{D}_{\kappa} &=& 
\left[
\begin{array}{cc}
\left( K_1 + \frac{n\kappa}{2}I\right)^2 & 0 \\
0 & \left( K_2 + \frac{n\kappa}{2}I\right)^2
\end{array}
\right].
\end{eqnarray}
As the bandwidth $\sigma$ of Gaussian kernel tends to zero, the population counterpart of the estimator converges to the mutual information up to second order when it is expanded around distributions with two variables $y_1$ and $y_2$ being independent \cite{Bach02JMLR}. 
The determinants of the Gram matrices $K_1$ and $K_2$ can be efficiently computed by using the incomplete Cholesky decomposition to find their low-rank approximations of rank $M$ ($\ll$ $n$). 
In \cite{Bach02JMLR}, it was suggested that the positive constant $\kappa$ and the width of the Gaussian kernel $\sigma$ are set to $\kappa=2\times10^{-3}$, $\sigma=1/2$ for $n$ $>$ 1000 and $\kappa=2\times10^{-2}$, $\sigma=1$ for $n$ $\le$ 1000 due to some theoretical and computational considerations. 

In this paper, we use the kernel-based independence measure. We first evaluate pairwise independence between a variable and each of the residuals and next take the sum of the pairwise measures over the residuals. 
Let us denote by $U$ the set of the subscripts of variables $x_i$, {\it i.e.}, $U$$=$\{$1$, $\cdots$, $p$\}. 
We use the following statistic to evaluate independence between a variable $x_j$ and its residuals $r_i^{(j)} = x_i - \frac{{\rm cov}(x_i,x_j)}{{\rm var}(x_j)}x_j$ when $x_i$ is regressed on $x_j$: 
\begin{eqnarray}
T_{kernel}(x_j; U) &=& \sum_{i\in U, i \neq j} \widehat{MI}_{kernel}(x_j, r_i^{(j)}). \label{eq:T}
\end{eqnarray}
Many other nonparametric independence measures \cite{Gretton05ALT,Kraskov04} and more computationally simple measures that use a single nonlinear correlation \cite{Hyva97NIPS} have also been proposed. 
Any such proposed method of independence could potentially be used instead of the kernel-based measure in Eq.~(\ref{eq:T}). 

%%%%%%%%%%%%%%%%%%%%%%%%%%%%%%%%%%%%
\subsection{DirectLiNGAM algorithm}\label{sec:alg}
%%%%%%%%%%%%%%%%%%%%%%%%%%%%%%%%%%%%
We now propose a new direct algorithm called DirectLiNGAM to estimate a causal ordering and the connection strengths in the LiNGAM (\ref{eq:model}):

\noindent
  %\pagebreak
  \rule{\columnwidth}{0.5mm}
       { \sffamily
	DirectLiNGAM algorithm
	 \begin{enumerate}
	 \item Given a $p$-dimensional random vector $\bfx$, a set of its variable subscripts $U$ and a $p \times n$ data matrix of the random vector as $\X$, initialize an ordered list of variables $K:=\emptyset$ and $m:=1$. 
	 \item Repeat until $p$$-$$1$  subscripts are appended to $K$:
	 \begin{enumerate}
	 \item \label{step:2a} Perform least squares regressions of $x_i$ on $x_j$ for all $i \in U \backslash  K$ ($i \neq j$) and compute the residual vectors $\bfr^{(j)}$ and the residual data matrix $\R^{(j)}$ from the data matrix $\X$ for all $j \in U \backslash  K$. Find a variable $x_m$ that is most independent of its residuals:  	 
	\begin{eqnarray}
	x_m=\arg\min_{j \in U \backslash  K} T_{kernel}( x_j; U \backslash  K), 
	\end{eqnarray}
	where $T_{kernel}$ is the independence measure defined in Eq.~(\ref{eq:T}). 
	\item Append $m$ to the end of $K$. 		
	\item Let $\bfx:=\bfr^{(m)}$, $\X:=\R^{(m)}$.
	 \end{enumerate}
	 \item Append the remaining variable to the end of $K$.
	 \item \label{step:reg} Construct a strictly lower triangular matrix $\B$ by following the order in $K$, and estimate the connection strengths $b_{ij}$ by using some conventional covariance-based regression such as least squares and maximum likelihood approaches on the original random vector $\bfx$ and the original data matrix $\X$. We use least squares regression in this paper. 
	 \end{enumerate}
       } \vspace{-4mm}
\noindent \rule{\columnwidth}{0.5mm}

%%%%%%%%%%%%%%%%%%%%%%%%%%%%%%%%%%%%
\subsection{Computational complexity}\label{sec:comp}
%%%%%%%%%%%%%%%%%%%%%%%%%%%%%%%%%%%%
Here, we consider the computational complexity of DirectLiNGAM compared with the ICA-LiNGAM  with respect to sample size $n$ and number of variables $p$. 
A dominant part of DirectLiNGAM is to compute Eq.~(\ref{eq:T}) for each $x_j$ in Step~2(a). 
Since it requires $O(np^2M^2+p^3M^3)$ operations \cite{Bach02JMLR} in $p$$-$$1$ iterations, 
complexity of the step is $O(np^3M^2+p^4M^3)$, where $M$ ($\ll$ $n$) is the maximal rank found by the low-rank decomposition used in the kernel-based independence measure. 
Another dominant part is the regression to estimate the matrix $\B$ in Step~\ref{step:reg}. 
The complexity of many representative regressions  including the least square algorithm is $O(np^3)$. 
Hence, we have a total budget of $O(np^3M^2+p^4M^3)$. 
Meanwhile, the ICA-LiNGAM requires $O(p^4)$ time to find a causal order in Step~\ref{step:causalperm}. 
Complexity of an iteration in FastICA procedure at Step~\ref{step:ica} is known to be $O(np^2)$. 
Assuming a constant number $C$ of the iterations in FastICA steps, the complexity of the ICA-LiNGAM is considered to be $O(Cnp^2+p^4)$. 
Though general evaluation of  the required iteration number $C$ is difficult, it can be conjectured to grow linearly with regards to $p$. Hence the complexity of the ICA-LiNGAM is presumed to be $O(np^3+p^4)$. 

Thus, the computational cost of DirectLiNGAM would be larger than that of ICA-LiNGAM especially when the low-rank approximation of the Gram matrices is not so efficient, {\it i.e.}, $M$ is large. 
However, we note the fact that DirectLiNGAM has guaranteed convergence in a fixed number of steps and is of known complexity, whereas for typical ICA algorithms including FastICA, the run-time complexity and the very convergence are not guaranteed.

%%%%%%%%%%%%%%%%%%%%%%%%%%%%%%%%%%%%
\subsection{Use of prior knowledge}\label{sec:prior}
%%%%%%%%%%%%%%%%%%%%%%%%%%%%%%%%%%%%
Although DirectLiNGAM requires no prior knowledge on the structure, 
more efficient learning can be achieved if some prior knowledge on a part of the structure is available because then the number of causal orders and connection strengths to be estimated gets smaller.  

We present three lemmas to utilize prior knowledge in DirectLiNGAM. 
Let us first define a matrix $\A^{{\rm knw}}$$=$$[a_{ji}^{{\rm knw}}]$ that collects prior knowledge under the LiNGAM (\ref{eq:model}) as follows:
\begin{eqnarray}
a_{ji}^{{\rm knw}}&:=&\left\{ 
\begin{array}{rl}
0 & \ {\rm if} \ x_i\ {\rm does\ {\it not}\ have\ a\ directed\ path\ to\ }x_j\\
1 & \ {\rm if} \ x_i\ {\rm has\ a\ directed\ path\ to\ }x_j\\
-1 & \ {\rm if \ no\ prior\ knowledge\ is\ available\ to\ know\ if\ either}\\
 & \ {\rm \ of \ the\ two\ cases\ above\ (0\ or\ 1)\ is\ true}.
\end{array}
\right.
\end{eqnarray}

Due to the definition of exogenous variables and that of prior knowledge matrix $\A^{{\rm knw}}$, we readily obtain the following three lemmas. 
\begin{lem}\label{lemma3}
Assume that the input data $\bfx$ strictly follows the LiNGAM (\ref{eq:model}). 
An observed variable $x_j$ is exogenous if $a_{ji}^{{\rm knw}}$ is zero for all $i$$\neq$$j$.
\end{lem}

\begin{lem}\label{lemma4}
Assume that the input data $\bfx$ strictly follows the LiNGAM (\ref{eq:model}). 
An observed variable $x_j$ is endogenous, {\it i.e.}, not exogenous, if there exist such $i$$\neq$$j$ that $a_{ji}^{{\rm knw}}$ is unity.
\end{lem}

\begin{lem}\label{lemma5}
Assume that the input data $\bfx$ strictly follows the LiNGAM (\ref{eq:model}). 
An observed variable $x_j$ does not receive the effect of $x_i$ if $a_{ji}^{{\rm knw}}$ is zero.
\end{lem}

The principle of making DirectLiNGAM algorithm more accurate and faster based on prior knowledge is as follows. 
We first find an exogenous variable by applying Lemma~\ref{lemma3} instead of Lemma~\ref{lemma1} if an exogenous variable is identified based on prior knowledge. 
Then we do not have to evaluate independence between any observed variable and its residuals. 
%Then, we do not have to compute residuals in linear regression or evaluate independence between any observed variable and its residuals. 
If no exogenous variable is identified based on prior knowledge, 
we next find endogenous (non-exogenous) variables by applying Lemma~\ref{lemma4}. 
Since endogenous variables are never exogenous we can narrow down the search space to find an exogenous variable based on Lemma~\ref{lemma1}. 
We can further skip to compute the residual of an observed variable and take the variable itself as the residual if its regressor does not receive the effect of the variable due to Lemma~\ref{lemma5}. 
Thus, we can decrease the number of causal orders and connection strengths to be estimated, and it improves the accuracy and computational time.  % as empirically demonstrated in Section~\ref{sec:exp}.  
The principle can also be used to further analyze the residuals and find the next exogenous residual because of Corollary~\ref{corollary3}. 
To implement these ideas, we only have to replace Step~\ref{step:2a} in DirectLiNGAM algorithm by the following steps: 

\noindent
       { \sffamily
	 \begin{enumerate}
	 \item[2a-1] Find such a variable(s) $x_j$ ($ j \in U \backslash  K$) that the $j$-th row of $\A^{{\rm knw}}$ has zero in the $i$-th column for all $i \in U \backslash  K$ ($i \neq j$) and denote the set of such variables by $U_{exo}$. If $U_{exo}$ is not empty, set $U_c := U_{exo}$. If $U_{exo}$ is empty, find such a variable(s) $x_j$ ($j \in U \backslash  K$) that the $j$-th row of $\A^{{\rm knw}}$ has unity in the $i$-th column for at least one of $i \in U \backslash  K$ ($i \neq j$), denote the set of such variables by $U_{end}$ and set $U_c := U \backslash  K \backslash  U_{end}$.  
	 \item[2a-2] Denote by $V^{(j)}$ a set of such a variable subscript $i \in U \backslash  K$ ($i \neq j$) that $a_{ij}^{{\rm  knw}} = 0$ for all $j \in U_c$.
         First set $\bfr_i^{(j)} := \bfx_i$ for all $i \in V^{(j)}$, next perform least squares regressions of $x_i$ on $x_j$ for all $i \in U \backslash  K \backslash  V^{(j)}$ ($i \neq j$) and estimate the residual vectors $\bfr^{(j)}$ and the residual data matrix $\R^{(j)}$ from the data matrix $\X$ for all $j \in U_c$. 
	 If $U_c$ has a single variable, set the variable to be $x_m$. 
	 Otherwise, find a variable $x_m$ in $U_c$ that is most independent of the residuals:  	 
	\begin{eqnarray}
	x_m=\arg\min_{j \in U_c} T_{kernel}( x_j; U\backslash K), 
	\end{eqnarray}
	where $T_{kernel}$ is the independence measure defined in Eq.~(\ref{eq:T}). 
	 \end{enumerate}
       } 

%%%%%%%%%%%%%%%%%%%%%%%%%%%%%%%%%%%%
\section{Simulations}\label{sec:exp}
%%%%%%%%%%%%%%%%%%%%%%%%%%%%%%%%%%%%

We first randomly generated 5 datasets based on sparse networks under each combination of number of variables $p$ and sample size $n$ ($p$$=$10, 20, 50, 100; $n$$=$500, 1000, 2000):
\begin{enumerate}
\item We constructed the $p \times p$ adjacency matrix with all zeros and replaced every element in the lower-triangular part by independent realizations of Bernoulli random variables with success probability $s$ similarly to \cite{Kalisch07JMLR}. The probability $s$ determines the sparseness of the model. The expected number of adjacent variables of each variable is given by $s(p-1)$. We randomly set the sparseness $s$ so that the number of adjacent variables was 2 or 5 \cite{Kalisch07JMLR}. 
\item We replaced each non-zero (unity) entry in the adjacency matrix by a value randomly chosen from the interval $[-1.5, -0.5]$ $\cup$ $[0.5, 1.5]$ and selected variances of the external influences $e_i$ from the interval $[1, 3]$ as in \cite{Silva06JMLR}. We used the resulting matrix as the data-generating adjacency matrix $\B$. 

\item \label{step:2} We generated data with sample size $n$ by independently drawing the external influence variables $e_i$ from various 18 non-Gaussian distributions used in \cite{Bach02JMLR} including (a) Student with 3 degrees of freedom; (b) double exponential; (c) uniform; (d) Student with 5 degrees of freedom; (e) exponential; (f) mixture of two double exponentials; (g)-(h)-(i) symmetric mixtures of two Gaussians: multimodal, transitional and unimodal; (j)-(k)-(l) nonsymmetric mixtures of two Gaussians, multimodal, transitional and unimodal; (m)-(n)-(o) symmetric mixtures of four Gaussians: multimodal, transitional and unimodal; (p)-(q)-(r) nonsymmetric mixtures of four Gaussians: multimodal, transitional and unimodal. See  Fig.~5 of \cite{Bach02JMLR} for the shapes of the probability density functions. 

\item The values of the observed variables $x_i$ were generated according to the LiNGAM (\ref{eq:model}). 
Finally, we randomly permuted the order of $x_i$. 
\end{enumerate}
Further we similarly generated 5 datasets based on dense (full) networks, {\it i.e.}, full DAGs with every pair of variables is connected by a directed edge, under each combination of number of variables $p$ and sample size $n$. 
Then we tested DirectLiNGAM and ICA-LiNGAM on the datasets generated by sparse networks or dense (full) networks. 
For ICA-LiNGAM, the maximum number of iterations was taken as 1000 \cite{Shimizu06JMLR}.  
The experiments were conducted on a standard PC using Matlab 7.9. Matlab implementations of the two methods are available on the web: \\
DirectLiNGAM: \url{http://www.ar.sanken.osaka-u.ac.jp/~inazumi/dlingam.html}\\
ICA-LiNGAM: \url{http://www.cs.helsinki.fi/group/neuroinf/lingam/}.

\begin{table}[!t] 
\caption{Median distances (Frobenius norms) between true $\B$ and estimated $\B$ of DirectLiNGAM and ICA-LiNGAM with five replications. } 
\label{tab:distance} 
\begin{center} 
\begin{tabular}{lr|rrr} 
\multicolumn{1}{l}{{\it Sparse} networks}  & & \multicolumn{3}{c}{Sample size} \\ 
\multicolumn{1}{c}{}  & &500 & 1000 & 2000\\ 
\hline
\hline
DirectLiNGAM & dim. = \hspace{1.75mm}10 & 0.48 & 0.31 & 0.21  \\
 & dim. = \hspace{1.75mm}20 &  1.19  & 0.70 & 0.50 \\
 & dim. = \hspace{1.75mm}50 & 2.57  & 1.82 & 1.40  \\
 & dim. = 100 & 5.75 & 4.61  & 2.35  \\
\hline 
ICA-LiNGAM & dim. = \hspace{1.75mm}10 & 3.01 & 0.74 & 0.65  \\ 
 & dim. = \hspace{1.75mm}20 &  9.68 &  3.00 & 2.06 \\
 & dim. = \hspace{1.75mm}50 & 20.61 & 20.23 & 12.91 \\
 & dim. = 100 &    40.77 & 43.74 & 36.52 \\
\hline
DirectLiNGAM with  & dim. = \hspace{1.75mm}10 &  0.48 & 0.30 & 0.24 \\
prior knowledge (50\%) & dim. = \hspace{1.75mm}20 & 1.00 & 0.71 & 0.49 \\
 & dim. = \hspace{1.75mm}50 & 2.47 & 1.75 & 1.19 \\
 & dim. = 100 & 4.94 & 3.89 & 2.27  \\
\multicolumn{5}{c}{} \\ 
\multicolumn{1}{c}{{\it Dense} (full) networks}  & & \multicolumn{3}{c}{Sample size} \\ 
\multicolumn{1}{c}{}  & &500 & 1000 & 2000\\ 
\hline
\hline
DirectLiNGAM & dim. = \hspace{1.75mm}10 & 0.45 	& 0.46 & 0.20  \\
 & dim. = \hspace{1.75mm}20 & 1.46	& 1.53 & 1.12  \\
 & dim. = \hspace{1.75mm}50 & 4.40 & 4.57 & 3.86  \\
 & dim. = 100 & 7.38 & 6.81 & 6.19  \\
\hline
 ICA-LiNGAM & dim. = \hspace{1.75mm}10 & 1.71 & 2.08 & 0.39  \\ 
 & dim. = \hspace{1.75mm}20  &   6.70 & 3.38 & 1.88  \\
 & dim. = \hspace{1.75mm}50 &     17.28  & 16.66 & 12.05  \\
 & dim. = 100 &    34.95 & 34.02 & 32.02  \\
\hline
DirectLiNGAM with  & dim. = \hspace{1.75mm}10 & 0.45  & 0.31 & 0.19  \\
prior knowledge (50\%) & dim. = \hspace{1.75mm}20 & 0.84 & 0.90 & 0.41 \\
 & dim. = \hspace{1.75mm}50 & 2.48  & 1.86 & 1.56 \\
 & dim. = 100 & 4.67  & 3.60 & 2.61 
\end{tabular} 
\end{center} 
\end{table} 

\begin{table}[!t] 
\caption{Median computational times (CPU times) of DirectLiNGAM and ICA-LiNGAM with five replications.} 
\label{tab:time} 
\begin{center} 
\begin{tabular}{lr|rrr} 
\multicolumn{1}{l}{{\it Sparse} networks}  & & \multicolumn{3}{c}{Sample size} \\ 
\multicolumn{1}{c}{}  & & \multicolumn{1}{c}{500} & \multicolumn{1}{c}{1000} & \multicolumn{1}{c}{2000}\\ 
\hline
\hline
DirectLiNGAM & dim. = \hspace{1.75mm}10 & 15.16 sec. & 37.21 sec. & 66.75 sec. \\
 & dim. = \hspace{1.75mm}20 &  1.56 {\small min.} & 5.75 {\small min.}& 17.22 {\small min.}\\
 & dim. = \hspace{1.75mm}50 & 16.25 {\small min.}& 1.34 hrs. & 2.70 hrs. \\
 & dim. = 100 & 2.35 hrs. & 21.17 hrs. & 19.90 hrs.\\
\hline
 ICA-LiNGAM & dim. = \hspace{1.75mm}10 &    0.73 sec.  & 0.41 sec.& 0.28 sec. \\ 
 & dim. = \hspace{1.75mm}20 &    5.40 sec. &   2.45 sec. & 1.14 sec.\\
 & dim. = \hspace{1.75mm}50 &    14.49 sec. &   21.47 sec. & 32.03 sec.\\
 & dim. = 100 &    46.32 sec. &  58.02 sec. & 1.16 {\small min.}\\
\hline
DirectLiNGAM with & dim. =\hspace{2.9mm}10 & 4.13 sec. &17.75 sec. & 30.95 sec. \\
 prior knowledge (50\%)  & dim. = \hspace{1.75mm}20 &  28.02 sec. & 1.64 {\small min.}& 4.98 {\small min.}\\
 & dim. =\hspace{1.75mm} 50 & 7.62 {\small min.}& 28.89 {\small min.}& 1.09 hrs. \\
 & dim. = 100 & 48.28 {\small min.}& 1.84 hrs. & 7.51 hrs.  \\
\multicolumn{5}{c}{} \\ 
\multicolumn{1}{c}{{\it Dense} (full) networks}  & & \multicolumn{3}{c}{Sample size} \\ 
\multicolumn{1}{c}{}  & & \multicolumn{1}{c}{500} & \multicolumn{1}{c}{1000} & \multicolumn{1}{c}{2000}\\ 
\hline
\hline
DirectLiNGAM & dim. = \hspace{1.75mm}10 & 8.05 sec. & 24.52 sec. & 49.44 sec.\\
 & dim. = \hspace{1.75mm}20 &  1.00 {\small min.} & 4.23 {\small min.}& 6.91 {\small min.}\\
 & dim. = \hspace{1.75mm}50 & 16.18 {\small min.}& 1.12 hrs. & 1.92 hrs. \\
 & dim. = 100 & 2.16 hrs. & 8.59 hrs. & 17.24 hrs. \\
\hline
 ICA-LiNGAM & dim. = \hspace{1.75mm}10 &       0.97 sec.  &  0.34 sec. &  0.27 sec. \\ 
 & dim. = \hspace{1.75mm}20  & 5.35 sec. & 1.25 sec.  & 4.07 sec.\\
 & dim. = \hspace{1.75mm}50 &     15.58 sec.& 21.01 sec.&  31.57 sec.\\
 & dim. = 100 &       47.60 sec. & 56.57 sec.&  1.36 {\small min.}\\
\hline
DirectLiNGAM with & dim. = \hspace{1.75mm}10 & 2.67 sec.& 5.66 sec.& 12.31 sec.\\
prior knowledge (50\%)  & dim. = \hspace{1.75mm}20 & 5.02 sec.& 31.70 sec. & 38.35 sec.\\
 & dim. = \hspace{1.75mm}50 & 46.74 sec. & 2.89 {\small min.}& 5.00 {\small min.}\\
 & dim. = 100 & 3.19 {\small min.}& 10.44 {\small min.}& 19.80 {\small min.}
\end{tabular} 
\end{center} 
\end{table} 

 \begin{figure*}[!tb]
\begin{center}
\includegraphics[width=3.25in]{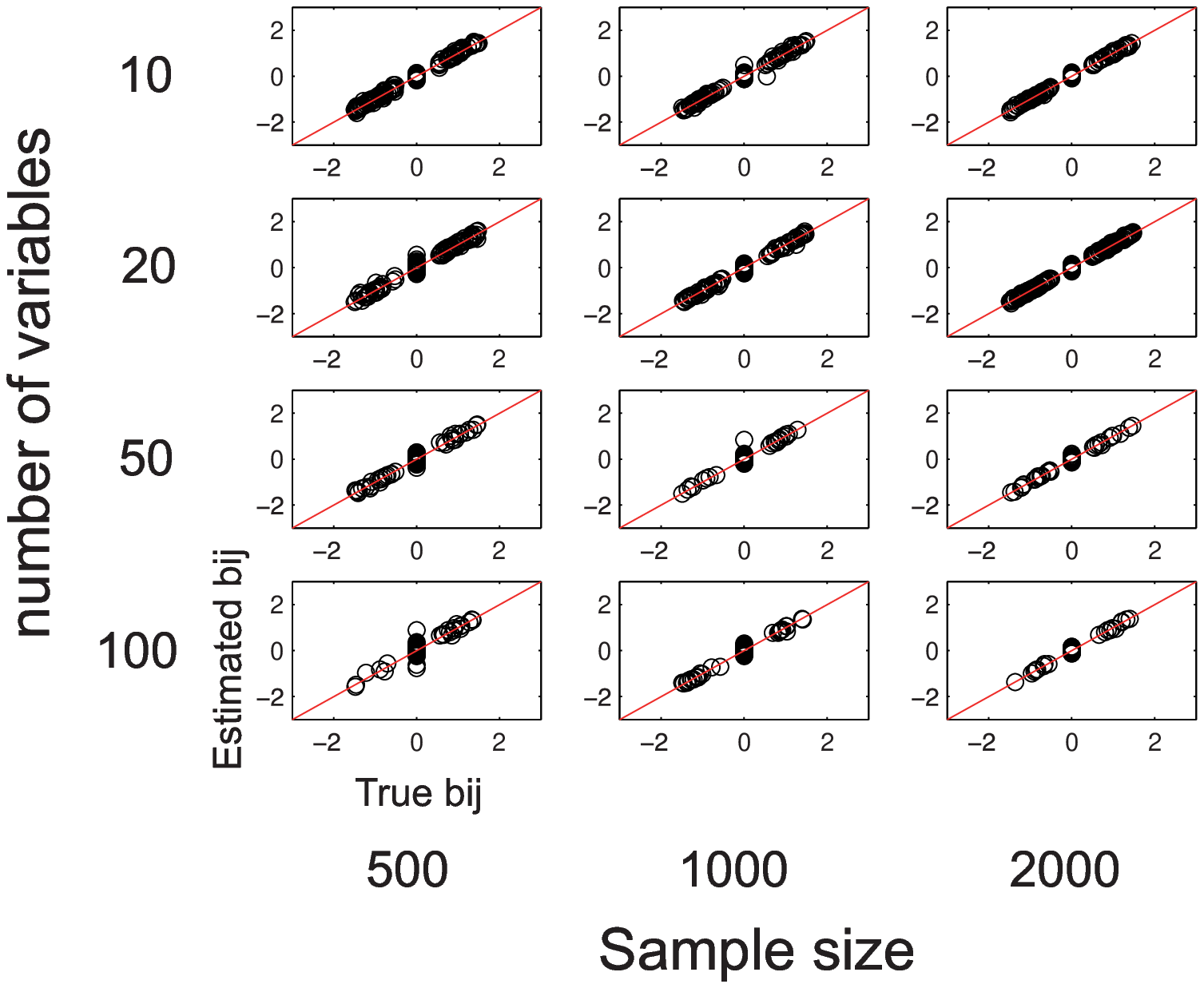}
\end{center}
\vspace{-4mm}
\caption{Scatterplots of the estimated $b_{ij}$ by DirectLiNGAM versus the true values for {\it sparse} networks.}
\label{fig:plots_DirectLiNGAM_B_sparse}
\end{figure*}

\begin{figure*}[!tb]
\begin{center}
\includegraphics[width=3.25in]{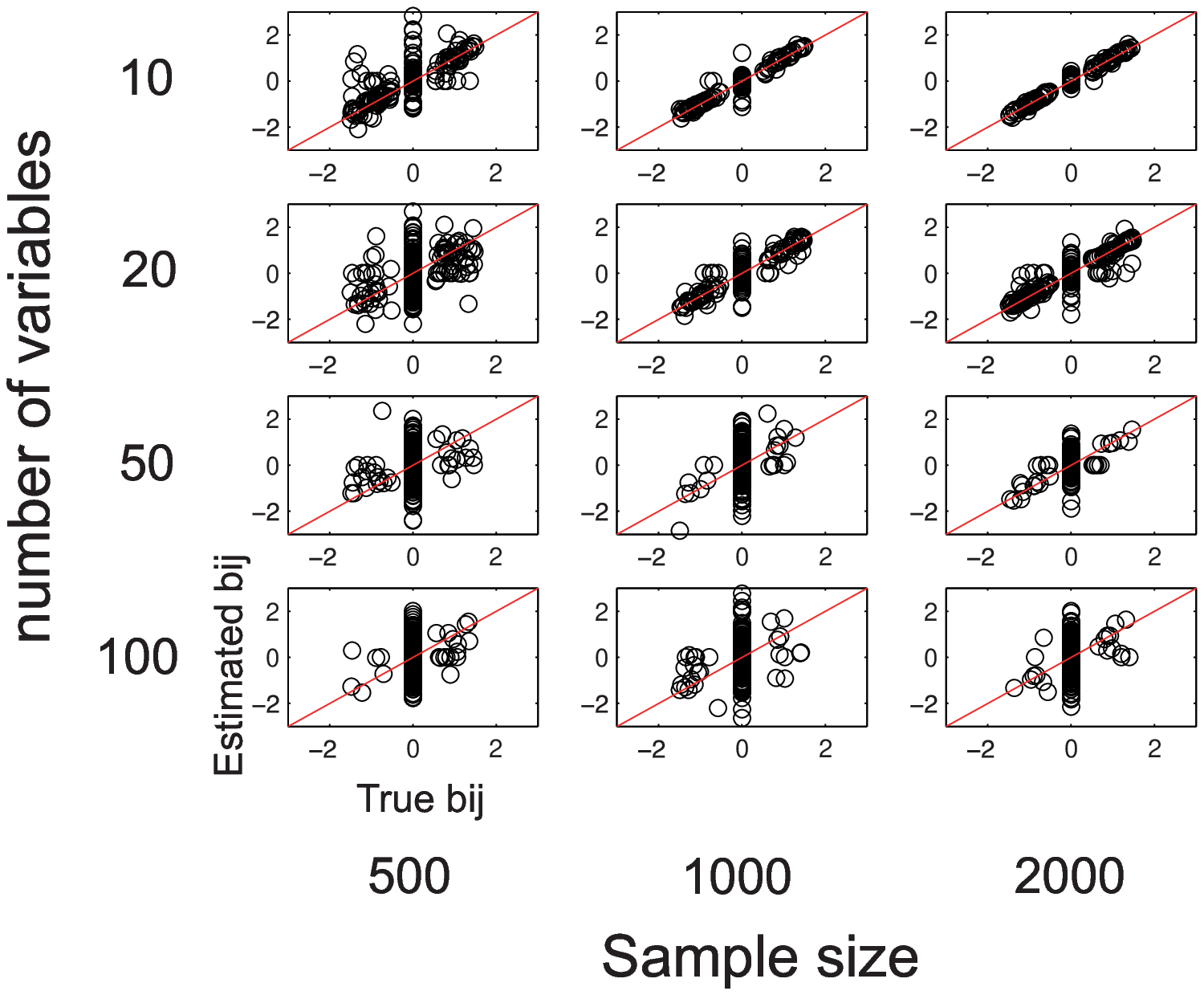}
\end{center}
\vspace{-4mm}
\caption{Scatterplots of the estimated $b_{ij}$ by ICA-LiNGAM versus the true values for {\it sparse} networks.}
\label{fig:plots_ICALiNGAM_B_sparse}
\end{figure*}

 \begin{figure*}[!tb]
\begin{center}
\includegraphics[width=3.25in]{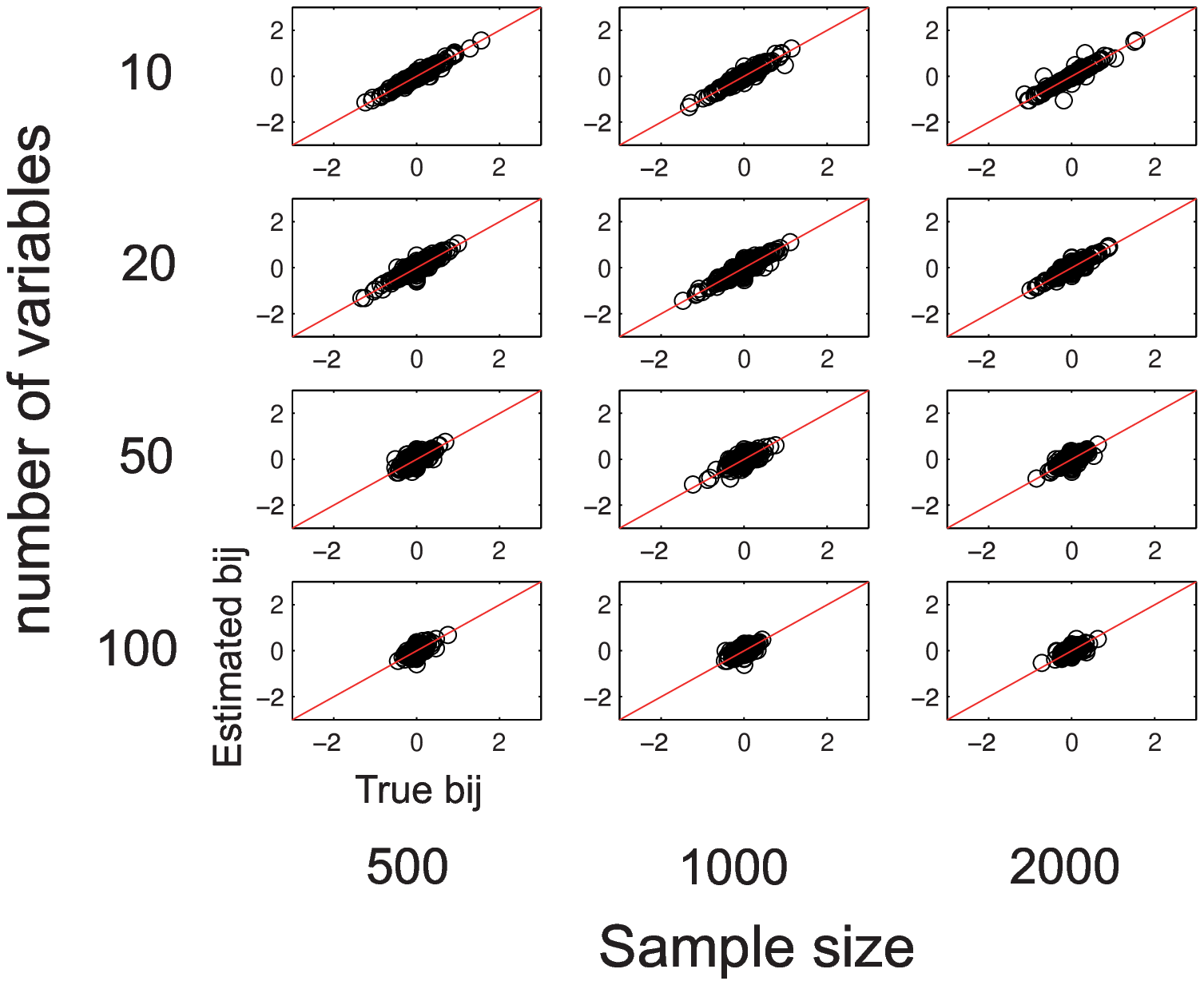}
\end{center}
\vspace{-4mm}
\caption{Scatterplots of the estimated $b_{ij}$ by DirectLiNGAM versus the true values for {\it dense} (full) networks.}
\label{fig:plots_DirectLiNGAM_B_dense}
\end{figure*}

\begin{figure*}[!tb]
\begin{center}
\includegraphics[width=3.25in]{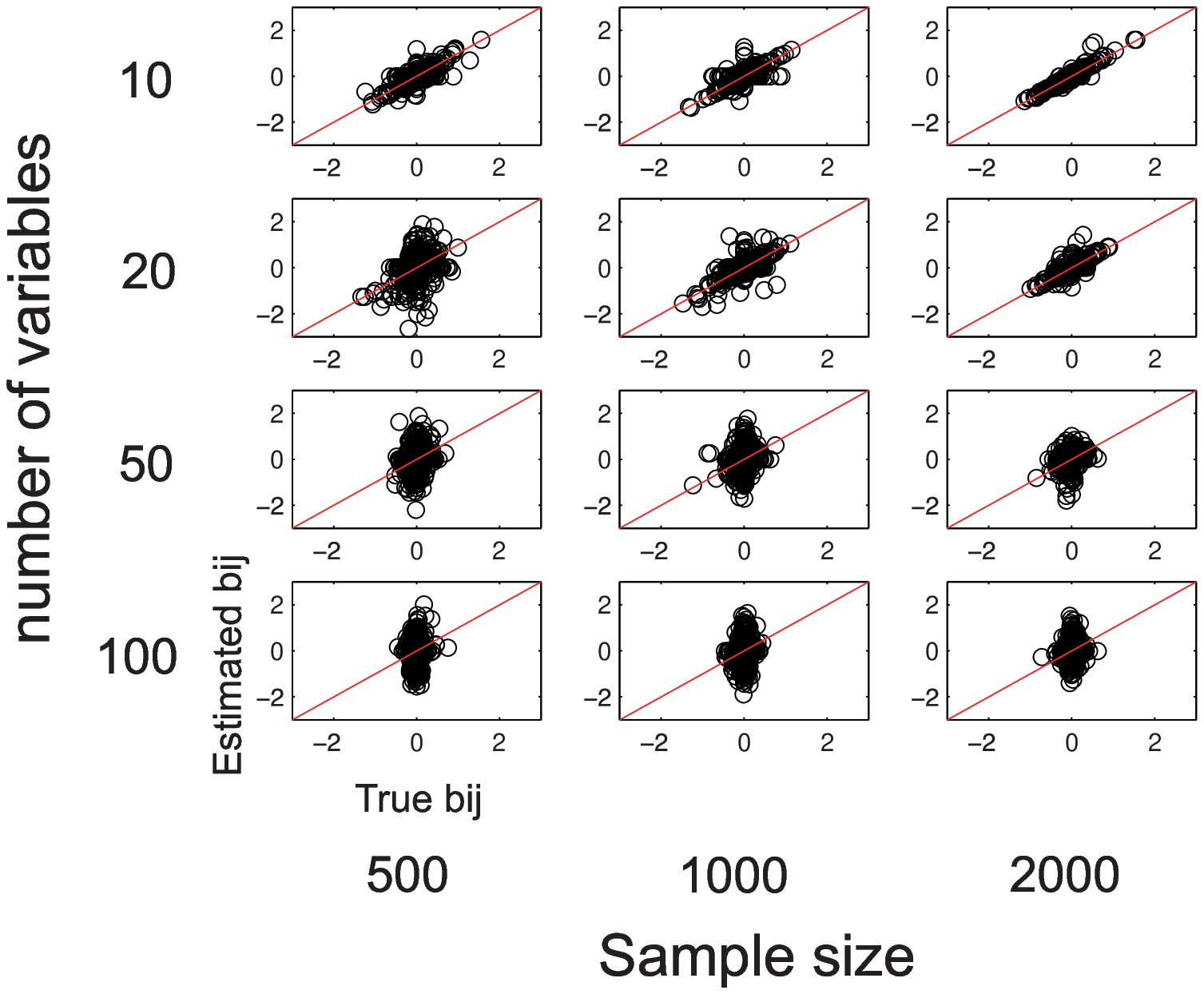}
\end{center}
\vspace{-4mm}
\caption{Scatterplots of the estimated $b_{ij}$ by ICA-LiNGAM versus the true values for {\it dense} (full) networks.}
\label{fig:plots_ICALiNGAM_B_dense}
\end{figure*}

 We computed the distance between the true $\B$ and  ones estimated by DirectLiNGAM and ICA-LiNGAM using the Frobenius norm defined as 
 \begin{eqnarray}
\sqrt{{\rm trace}\{(\B_{true}-\widehat{\B})^T(\B_{true}-\widehat{\B})\}}.
 \end{eqnarray}
Tables~\ref{tab:distance} and \ref{tab:time} show the median distances (Frobenius norms) and median computational times  (CPU times) respectively. 
In Table~\ref{tab:distance}, DirectLiNGAM was better in distances of $\B$ and gave more accurate estimates of $\B$ than ICA-LiNGAM for all of the conditions. 
In Table~\ref{tab:time}, the computation amount of DirectLiNGAM was rather larger than ICA-LiNGAM when the sample size was increased. A main bottleneck of computation was the kernel-based independence measure. 
However, its computation amount can be considered to be still tractable. 
In fact, the actual elapsed times were approximately one-quarter of their CPU times respectively probably because the CPU had four cores. 
Interestingly, the CPU time of ICA-LiNGAM actually decreased with increased sample size in some cases. This is presumably due to better convergence properties. 

 To visualize the estimation results, Figures~\ref{fig:plots_DirectLiNGAM_B_sparse}, \ref{fig:plots_ICALiNGAM_B_sparse}, \ref{fig:plots_DirectLiNGAM_B_dense} and \ref{fig:plots_ICALiNGAM_B_dense} give combined scatterplots of the estimated elements of $\B$ of DirectLiNGAM and ICA-LiNGAM versus the true ones for sparse networks and dense (full) networks respectively. The different plots correspond to different numbers of variables and different sample sizes, where each plot combines the data for different adjacency matrices $\B$ and 18 different distributions of the external influences $p(e_i)$. 
We can see that DirectLiNGAM worked well and better than ICA-LiNGAM, as evidenced by the grouping of the data points onto the main diagonal.

Finally, we generated datasets in the same manner as above and gave some prior knowledge to DirectLiNGAM
 by creating prior knowledge matrices $\A^{{\rm knw}}$ as follows. 
We first replaced every non-zero element by unity and every diagonal element by zero in $\A$$=$$(\I-\	B)^{-1}$ and subsequently hid each of the off-diagonal elements, {\it i.e.}, replaced it by $-1$, with probability $0.5$. 
The bottoms of Tables~\ref{tab:distance} and \ref{tab:time} show the median distances and median computational times. 
It was empirically confirmed that use of prior knowledge gave more accurate estimates and less computational times in most cases especially for dense (full) networks. 
The reason would probably be that for dense (full) networks more prior knowledge about where directed paths exist were likely to be given and it narrowed down the search space more efficiently. 

%%%%%%%%%%%%%%%%%%%%%%%%%%%%%%%%%%%%
\section{Applications to real-world data}\label{sec:real}
%%%%%%%%%%%%%%%%%%%%%%%%%%%%%%%%%%%%
We here apply DirectLiNGAM and ICA-LiNGAM on real-world physics and sociology data. Both DirectLiNGAM and ICA-LiNGAM estimate a causal ordering of variables and provide a full DAG. Then we have two options to do further analysis \cite{Hyva10JMLR}: i) Find significant directed edges or direct causal effects $b_{ij}$ and significant total causal effects $a_{ij}$ with $\A$$=$$(\I-\B)^{-1}$; ii) Estimate redundant directed edges to find the underlying DAG. We demonstrate an example of the former in Subsection~\ref{sec:real_furiko} and that of the latter in Subsection~\ref{sec:real_soc}.

%%%%%%%%%%%%%%%%%%%%%%%%%%%%%%%%%%%%
\subsection{Application to physical data}\label{sec:real_furiko}
\begin{figure*}[!tb]
\begin{center}
\includegraphics[width=2in]{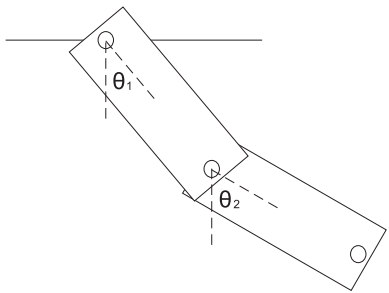}
\end{center}
\vspace{-4mm}
\caption{Abstract model of the double-pendulum used in \cite{Kawahara11ARMA}.}
\label{fig:furiko}
\end{figure*}

%%%%%%%%%%%%%%%%%%%%%%%%%%%%%%%%%%%%
We applied DirectLiNGAM and ICA-LiNGAM on a dataset created from a physical system called a double-pendulum, a pendulum with another pendulum attached to its end \cite{Meirovitch86book} as in Fig.~\ref{fig:furiko}. 
The dataset was first used in \cite{Kawahara11ARMA}. 
The raw data consisted of four time series provided by Ibaraki University (Japan) filming the pendulum system with a high-speed video camera at every 0.01 second for 20.3 seconds and then reading out the position using an image analysis software. 
The four variables were $\theta_1$: the angle between the top limb and the vertical, $\theta_2$: the angle between the bottom limb and the vertical, $\omega_1$: the angular speed of $\theta_1$ or $\dot{\theta}_1$ and $\omega_2$: the angular speed of $\theta_2$ or $\dot{\theta}_2$. The number of time points was 2035. 
The dataset is available on the web: \\\url{http://www.ar.sanken.osaka-u.ac.jp/~inazumi/data/furiko.html}

In \cite{Kawahara11ARMA}, some theoretical considerations based on the domain knowledge implied that the angle speeds $\omega_1$ and $\omega_2$ are mainly determined by the angles $\theta_1$ and $\theta_2$ in both cases where the swing of the pendulum is sufficiently small ($\theta_1, \theta_2 \approx 0$) and where the swing is not very small. Further, in practice, it was reasonable to assume that there were no latent confounders \cite{Kawahara11ARMA}. 

\begin{figure*}[!tb]
\begin{center}
\includegraphics[width=3.75in]{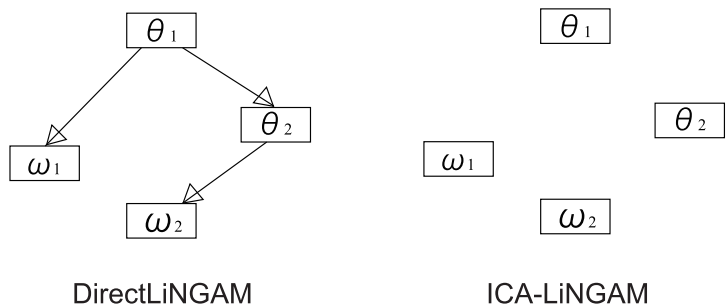}
\end{center}
\vspace{-4mm}
\caption{Left: The estimated network by DirectLiNGAM. Only significant directed edges are shown with 5\% significance level. Right: The estimated network by ICA-LiNGAM. No significant directed edges were found with 5\% significance level.}
\label{fig:furiko_DirectLiNGAM_ICALiNGAM}
\end{figure*}

\begin{figure*}[!tb]
\begin{center}
\includegraphics[width=3.75in]{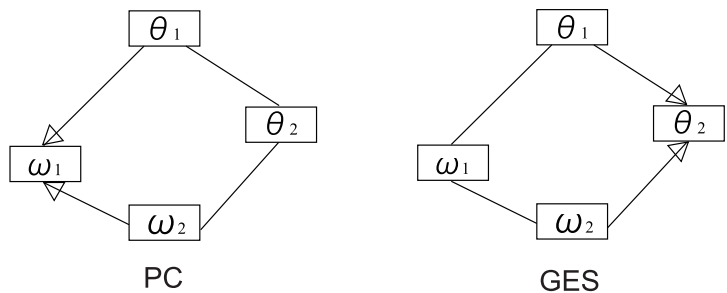}
\end{center}
\vspace{-4mm}
\caption{Left: The estimated network by PC algorithm with 5\% significance level. Right: The estimated network by GES. An undirected edge between two variables means that there is a directed edge from a variable to the other or the reverse. }
\label{fig:furiko_PC_GES}
\end{figure*}

As a preprocessing, we first removed the time dependency from the raw data using the ARMA (AutoRegressive Moving Average) model with 2 autoregressive terms and 5 moving average terms following \cite{Kawahara11ARMA}. 
 Then we applied DirectLiNGAM and ICA-LiNGAM on the preprocessed data. 
The estimated adjacency matrices $\B$ of $\theta_1$, $\theta_2$, $\omega_1$ and $\omega_2$ were as follows: 
\begin{eqnarray}
{\rm DirectLiNGAM}&:&
\bordermatrix{ 
  & \theta_1 & \theta_2 & \omega_1 & \omega_2 \cr
\theta_1  &  0 &  0 & 0 & 0 \cr
\theta_2 &  -0.23 & 0 & 0 & 0\cr
\omega_1 &  90.39  & -2.88 & 0 & 0\cr
\omega_2 &   5.65 &  94.64 &  -0.11 &  0 \cr
},\\            
{\rm ICA-LiNGAM}&:&
\bordermatrix{ 
  & \theta_1 & \theta_2 & \omega_1 & \omega_2 \cr
\theta_1  &  0 &  0 & 0 & 0 \cr
\theta_2 &  1.45 & 0 & 0 & 0\cr
\omega_1 &  108.82  & -52.73 & 0 & 0\cr
\omega_2 &    216.26  & 112.50 &  -1.89 & 0 \cr
}.
\end{eqnarray}
The estimated orderings by DirectLiNGAM and ICA-LiNGAM were identical, but the estimated connection strengths were very different. 
We further computed their 95\% confidence intervals by using bootstrapping \cite{Efron93book} with the number of bootstrap replicates 10000. 
The estimated networks by DirectLiNGAM and ICA-LiNGAM are graphically shown in Fig.~\ref{fig:furiko_DirectLiNGAM_ICALiNGAM}, where only significant directed edges (direct causal effects) $b_{ij}$ are shown with 5\% significance level.\footnote{The issue of multiple comparisons arises in this context, which we would like to study in future work.}
DirectLiNGAM found that the angle speeds $\omega_1$ and $\omega_2$ were determined by the angles $\theta_1$ or $\theta_2$, which was consistent with the domain knowledge. 
Though the directed edge from $\theta_1$ to $\theta_2$ might be a bit difficult to interpret, the effect of $\theta_1$ on $\theta_2$ was estimated to be negligible since the coefficient of determination \cite{Bollen89book} of $\theta_2$, {\it i.e.}, $1$$-${\rm var}($\hat{e}_2$)/{\rm var}($\hat{\theta}_2$), was very small and was 0.01. 
(The coefficient of determination of $\omega_1$ and that of $\omega_2$ were 0.46 and 0.49 respectively.)
On the other hand, ICA-LiNGAM could not find any significant directed edges since it gave very different estimates for different bootstrap samples. 

For further comparison, we also tested two conventional methods \cite{Spirtes91,Chic02JMLR} based on conditional independences. Fig.~\ref{fig:furiko_PC_GES} shows the estimated networks by PC algorithm \cite{Spirtes91} with 5\% significance level  and GES \cite{Chic02JMLR} with the Gaussianity assumption. 
We used the Tetrad IV\footnote{\url{http://www.phil.cmu.edu/projects/tetrad/}} to run the two methods. 
PC algorithm found the same directed edge from $\theta_1$ on $\omega_1$ as DirectLiNGAM did but  did not found the directed edge from $\theta_2$ on $\omega_2$. GES found the same directed edge from $\theta_1$ on $\theta_2$ as DirectLiNGAM did but did not find that the angle speeds $\omega_1$ and $\omega_2$ were determined by the angles $\theta_1$ or $\theta_2$.

We also computed the 95\% confidence intervals of the total causal effects $a_{ij}$ using bootstrap. 
DirectLiNGAM found significant total causal effects from $\theta_1$ on $\theta_2$, 
from $\theta_1$ on $\omega_1$, 
from $\theta_1$ on $\omega_2$, 
from $\theta_2$ on $\omega_1$,
and 
from $\theta_2$ on $\omega_2$. 
These significant total effects would also be reasonable based on similar arguments. 
ICA-LiNGAM only found a significant total causal effect
from $\theta_2$ on $\omega_2$. 

Overall, although the four variables $\theta_1$, $\theta_2$, $\omega_1$ and $\omega_2$ are likely to be nonlinearly related  according to the domain knowledge \cite{Meirovitch86book,Kawahara11ARMA}, DirectLiNGAM gave interesting results in this example. 

%%%%%%%%%%%%%%%%%%%%%%%%%%%%%%%%%%%%
\subsection{Application to sociology data}\label{sec:real_soc}
%%%%%%%%%%%%%%%%%%%%%%%%%%%%%%%%%%%%

\begin{figure*}[!tb]
\centering
\includegraphics[width=3.5in]{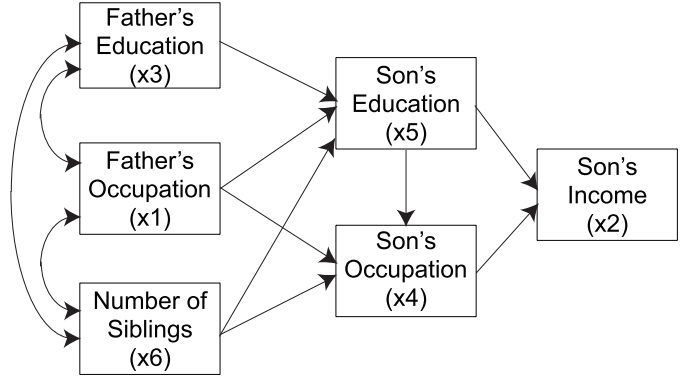}
\caption{Status attainment model based on domain knowledge \cite{Duncan72book}. A bi-directed edge between two variables means that the relation is not modeled. For instance, there could be latent confounders between the two, there could be a directed edge between the two, or the two could be independent.}
\label{fig:background}
\end{figure*}

We analyzed a dataset taken from a sociological data repository on the Internet called General Social Survey ({\url{http://www.norc.org/GSS+Website/}). 
The data consisted of six observed variables, $x_1$: father's occupation level, $x_2$: son's income, $x_3$: father's education, $x_4$: son's occupation level, $x_5$: son's education, $x_6$: number of siblings. 
The sample selection was conducted based on the following criteria: i) non-farm background (based on two measures of father's occupation); ii) ages 35 to 44; iii) white; iv) male; v) in the labor force at the time of the survey; vi) not missing data for any of the covariates; vii) years 1972-2006. The sample size was 1380. Fig.~\ref{fig:background} shows domain knowledge about their causal relations \cite{Duncan72book}. 
As shown in the figure, there could be some latent confounders between $x_1$ and $x_3$, $x_1$ and $x_6$, or $x_3$ and $x_6$. 
An objective of this example was to see how our method behaves when such a model assumption of LiNGAM could be violated that there is no latent confounder. 

The estimated adjacency matrices $\B$ by DirectLiNGAM and ICA-LiNGAM were as follows: 
\begin{eqnarray}
{\rm DirectLiNGAM: }
\bordermatrix{ 
  & x_1 & x_2 & x_3 & x_4 & x_5 & x_6 \cr
x_1 & 0  & 0  & 3.19  & 0.10  & 0.41  & 0.21 \cr
x_2 & 33.48  & 0  & 452.84 & 422.87  & 1645.45  & 347.96 \cr
x_3 &0  & 0  & 0  & 0  & 0.55  & -0.18 \cr
x_4 &0  & 0  & 0.17  & 0  & 4.61  & -0.19 \cr
x_5 &0  & 0  & 0  & 0  & 0  & -0.12 \cr
x_6 &0  & 0  & 0  & 0  & 0  & 0 \cr
},\\            
{\rm ICA-LiNGAM: } 
\bordermatrix{ 
  & x_1 & x_2 & x_3 & x_4 & x_5 & x_6 \cr
x_1 & 0  & 0  & 0.93  & 0  & -0.68  & -0.20 \cr
x_2 &50.70  & 0  & -31.82  & 200.84  & 65.63  & 336.04 \cr
x_3 &0  & 0  & 0  & 0  & 0.24  & -0.27 \cr
x_4 &0.17  & 0  & -0.40  & 0  & -0.14  & -0.14 \cr
x_5 &0  & 0  & 0  & 0  & 0  & 0 \cr
x_6 &0  & 0  & 0  & 0  & -0.08  & 0 
}.
\end{eqnarray}

We subsequently pruned redundant directed edges $b_{ij}$ in the full DAGs by repeatedly applying a sparse method called Adaptive Lasso \cite{Zou06JASA} on each variable and its potential parents.  
See Appendix A for some more details of Adaptive Lasso. 
We used a matlab implementation in \cite{IMM2005-03897} to run the Lasso. Then we obtained the following pruned adjacency matrices $\B$: 
\begin{eqnarray}
{\rm DirectLiNGAM} &:&
\bordermatrix{ 
  & x_1 & x_2 & x_3 & x_4 & x_5 & x_6 \cr
x_1 &0  & 0  & 3.19  & 0  & 0 & 0 \cr
x_2 &0  & 0  & 0 & 422.87  & 0  & 0 \cr
x_3 &0  & 0  & 0  & 0  & 0.55  & 0 \cr
x_4 &0  & 0  & 0 & 0  & 4.61  & 0 \cr
x_5 &0  & 0  & 0  & 0  & 0  & -0.12 \cr
x_6 &0  & 0  & 0  & 0  & 0  & 0 \cr
},\\            
{\rm ICA-LiNGAM}&:&
\bordermatrix{ 
  & x_1 & x_2 & x_3 & x_4 & x_5 & x_6 \cr
x_1 &0  & 0  & 0.93  & 0  & 0  & 0 \cr
x_2 &0 & 0  & 0  & 200.84  & 0  & 0 \cr
x_3 &0  & 0  & 0  & 0  & \hspace{3mm} 0.24  & 0 \cr
x_4 &0  & 0  & 0  & 0  & -0.14  & 0 \cr
x_5 &0  & 0  & 0  & 0  & 0  & 0 \cr
x_6 &0  & 0  & 0  & 0  & -0.08  & 0 \cr}.
\end{eqnarray}

The estimated networks by DirectLiNGAM and ICA-LiNGAM are graphically shown in Fig.~\ref{fig:directlingam_soc} and Fig.~\ref{fig:icalingam_soc} respectively. 
All the directed edges estimated by DirectLiNGAM were reasonable to the domain knowledge other than the directed edge from $x_5$: son's education to $x_3$: father's education. 
Since the sample size was large and yet the estimated model was not fully correct, the mistake on the directed edge between $x_5$ and $x_3$ might imply that some model assumptions might be more or less violated in the data. 
ICA-LiNGAM gave a similar estimated network but did one more mistake that $x_6$: number of siblings  is determined by $x_5$: son's education.

Further, Fig.~\ref{fig:PC_soc} and Fig.~\ref{fig:GES_soc} show the estimated networks by PC algorithm with 5\% significance level  and GES with the Gaussianity assumption. 
Both of the conventional methods did not find the directions of many edges. 
The two conventional methods found a reasonable direction of the edge between  $x_1$: father's occupation and $x_3$: father's education, but they gave a wrong direction of the edge between $x_1$: father's occupation and $x_4$: son's occupation. 

\begin{figure*}[!tb]
\centering
\includegraphics[width=3.25in]{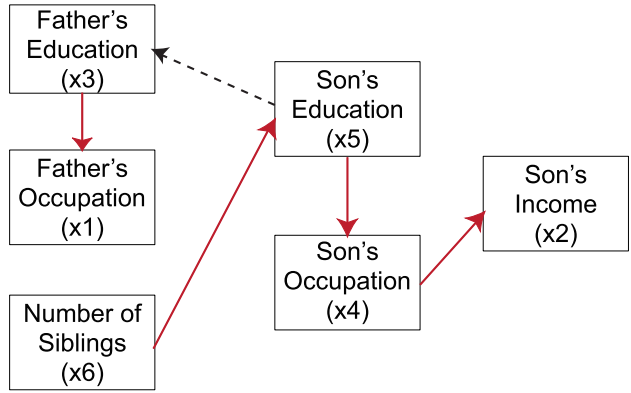}
\caption{The estimated network by DirectLiNGAM and Adaptive Lasso. A red solid directed edge is reasonable to the domain knowledge.}
\label{fig:directlingam_soc}
\end{figure*}

\begin{figure*}[!tb]
\centering
\includegraphics[width=3.25in]{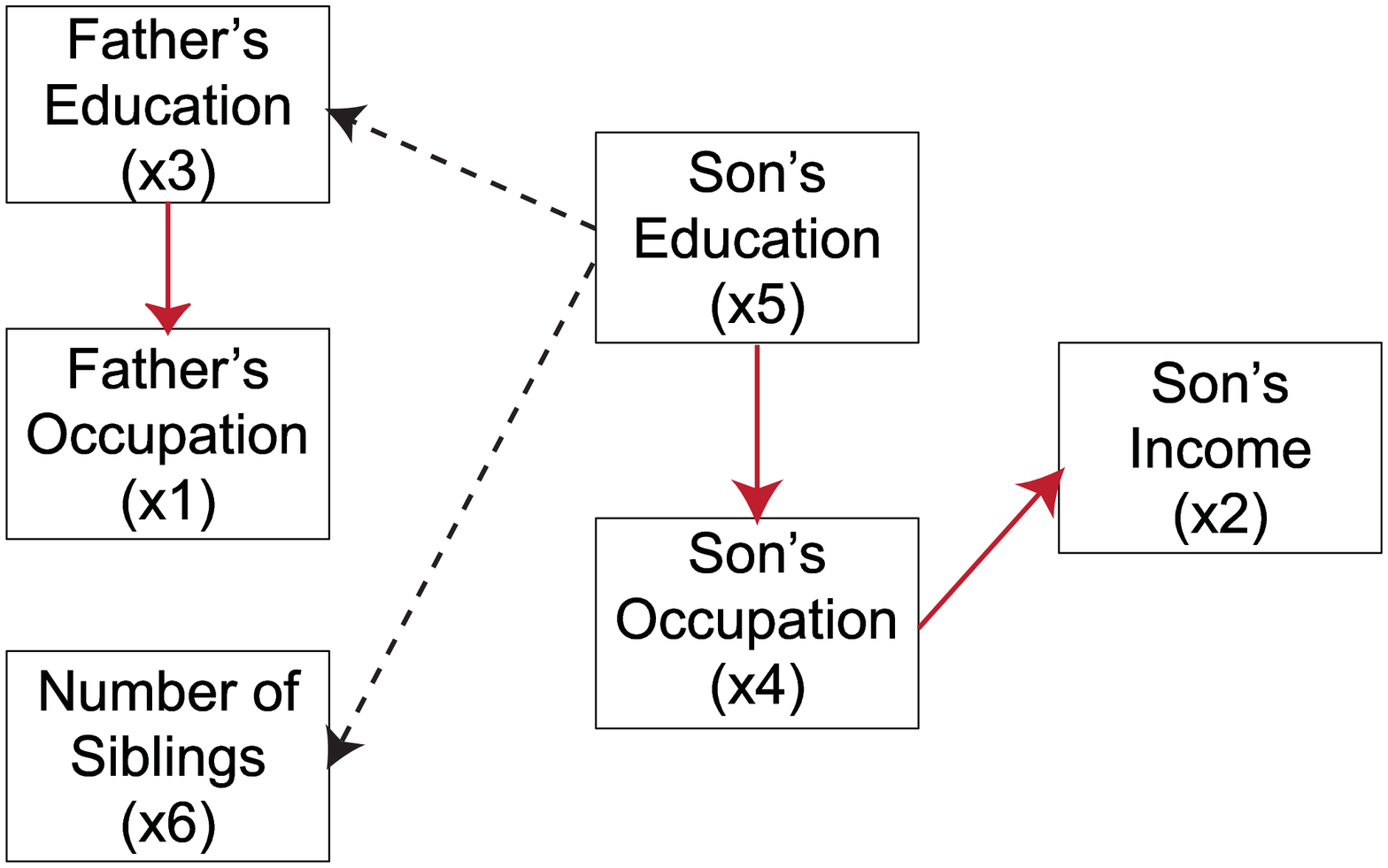}
\caption{The estimated network by ICA-LiNGAM and Adaptive Lasso. A red solid directed edge is reasonable to the domain knowledge.}
\label{fig:icalingam_soc}
\end{figure*}

\begin{figure*}[!tb]
\begin{center}
\includegraphics[width=3.5in]{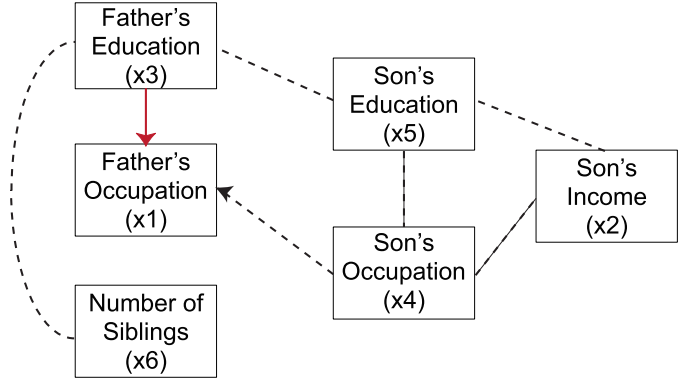}
\end{center}
\vspace{-4mm}
\caption{The estimated network by PC algorithm with 5\% significance level. An undirected edge between two variables means that there is a directed edge from a variable to the other or the reverse. A red solid directed edge is reasonable to the domain knowledge. }
\label{fig:PC_soc}
\end{figure*}

\begin{figure*}[!tb]
\begin{center}
\includegraphics[width=3.5in]{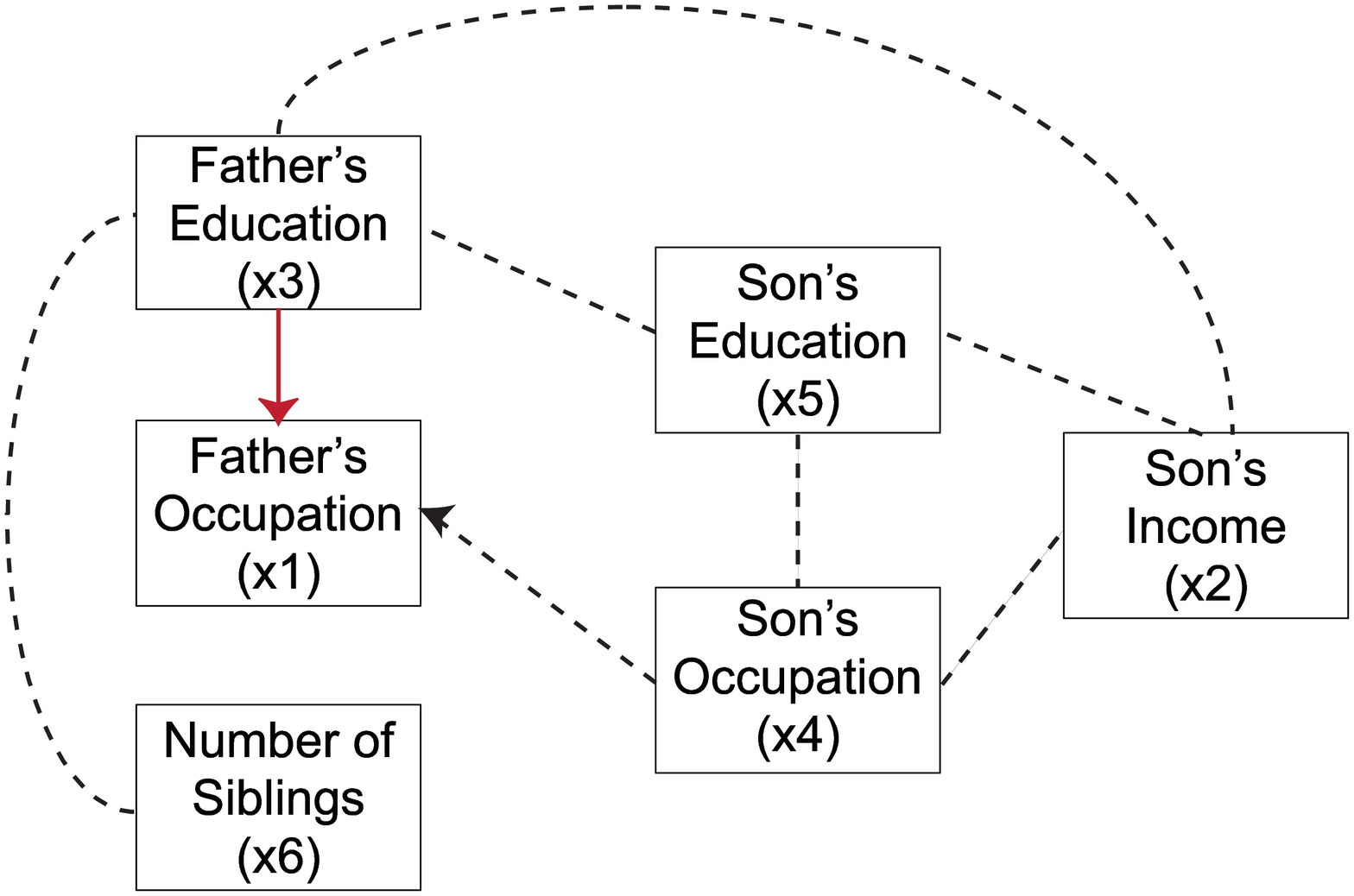}
\end{center}
\vspace{-4mm}
\caption{The estimated network by GES. An undirected edge between two variables means that there is a directed edge from a variable to the other or the reverse. A red solid directed edge is reasonable to the domain knowledge. }
\label{fig:GES_soc}
\end{figure*}

%%%%%%%%%%%%%%%%%%%%%%%%%%%%%%%%%%%%
\section{Conclusion}\label{sec:conc}
%%%%%%%%%%%%%%%%%%%%%%%%%%%%%%%%%%%%
We presented a new estimation algorithm for the LiNGAM that has guaranteed convergence  to the right solution in a fixed number of steps if the data strictly follows the model and known computational complexity unlike most ICA methods. 
This is the first algorithm specialized to estimate the LiNGAM. 
Simulations implied that the new method often provides better statistical performance than a state of the art method based on ICA. In real-world applications to physics and sociology, promising results were obtained. 
Future works would include i) assessment of practical performance of statistical tests to detect violations of the model assumptions including tests of independence \cite{Gretton10JMLR}; ii) implementation issues of our algorithm to improve the practical computational efficiency. 

\section*{Acknowledgements}
We are very grateful to Hiroshi Hasegawa (College of Science, Ibaraki University, Japan) for providing the physics data and Satoshi Hara and Ayumu Yamaoka for interesting discussion.  
This work was partially carried out at Department of Mathematical and Computing Sciences and Department of Computer Science, Tokyo Institute of Technology, Japan. 
S.S., Y.K. and T.W. were partially supported  by MEXT Grant-in-Aid for Young Scientists \#21700302, by JSPS Grant-in-Aid for Young Scientists \#20800019 and by Grant-in-Aid for Scientific Research (A) \#19200013, respectively. 
S.S. and Y.K. were partially supported by JSPS Global COE program
`Computationism as a Foundation for the Sciences'. 
A.H. was partially supported by the Academy of Finland Centre of Excellence for Algorithmic Data Analysis. 

%%%%%%%%%%%%%%%%%%%%%%%%%%%%%%%%%%%%%
\appendix

\section{Adaptive Lasso}\label{sec:lasso}
We very briefly review the adaptive Lasso \cite{Zou06JASA}, which is a variant of the Lasso \cite{Tibshiranilasso}. See \cite{Zou06JASA} for more details. 
The adaptive Lasso is a regularization technique for variable selection and assumes the same data generating process as LiNGAM: 
\begin{equation}
x_i = \sum_{k(j)<k(i)} b_{ij}x_j + e_i.\label{eq:lasso}
\end{equation}
A big difference is that the adaptive Lasso assumes that the set of such potential parent variables $x_j$ that $k(j)$$<$$k(i)$ is known and LiNGAM estimates the set of such variables. 
The adaptive Lasso penalizes connection strengths $b_{ij}$ in $L_1$ penalty by minimizing the objective function defined as: 
\begin{eqnarray}
\left\| x_i-\sum_{k(j)<k(i)} b_{ij}x_j \right\|^2 + \lambda \sum_{k(j)<k(i)} \frac{|b_{ij}|^{\hspace{1.5mm}}}{|\hat{b}_{ij}|^{\gamma}},
\end{eqnarray}
where $\lambda$ and $\gamma$ are tuning parameters and $\hat{b}_{ij}$ is a consistent estimate of $b_{ij}$. 
In \cite{Zou06JASA}, it was suggested to select the tuning parameters by five-fold cross validation and to obtain $\hat{b}_{ij}$ by ordinary least squares regression. 
The adaptive Lasso has a very attractive property that it asymptotically selects the right set of such variables $x_j$ that $b_{ij}$ is not zero, where $k(j)$$<$$k(i)$. 

% References
\bibliography{Dlingam_Ref}
\bibliographystyle{splncs}

\end{document}